\newif \ifusenix
\newif \ifacm
\newif \ifmcom
\newif \ifimwut

\usenixfalse
\acmfalse
\mcomfalse
\imwuttrue
\newcommand{\name}{WiForceSticker\xspace}

\ifusenix
	\documentclass[letterpaper,twocolumn,9pt]{article}
	\usepackage{usenix}
	\usepackage{times}
\fi

\ifacm
\documentclass[sigconf,10pt]{acmart}

\usepackage[english]{babel}
\usepackage{blindtext}

\renewcommand\footnotetextcopyrightpermission[1]{} 
\setcopyright{none}

\acmDOI{}

\acmISBN{}

\acmConference[Submitted for review to SenSys]{}
\acmYear{2020}

\acmPrice{}
\fi

\ifimwut
\documentclass[acmlarge]{acmart}

\usepackage[english]{babel}
\usepackage{blindtext}

\AtBeginDocument{%
  \providecommand\BibTeX{{%
    \normalfont B\kern-0.5em{\scshape i\kern-0.25em b}\kern-0.8em\TeX}}}

\renewcommand\footnotetextcopyrightpermission[1]{} 
\setcopyright{none}

\copyrightyear{2021}
\acmYear{2021}
\acmDOI{0}

\acmJournal{IMWUT}
\acmVolume{0}
\acmNumber{0}
\acmArticle{0}
\acmMonth{0}
\fi

\ifmcom
  \documentclass[sigconf]{sig-alternate-10pt}
\fi
\usepackage{xcolor}
\usepackage{amsfonts}
\PassOptionsToPackage{hyphens}{url}\usepackage{hyperref}
\usepackage{color}
\usepackage{graphics}
\usepackage{graphicx}
\usepackage{url}
\usepackage{listings}
\usepackage{multicol}
\usepackage{multirow}
\usepackage[scaled]{helvet}
\usepackage{rotating}
\usepackage{xspace}
\usepackage[ruled,vlined]{algorithm2e}
\usepackage{comment}
\usepackage{enumitem}
\usepackage{amsmath}
\usepackage{mathrsfs}
\usepackage{cancel}
\usepackage{cleveref}
\usepackage{subcaption}
\usepackage{float}
\usepackage[font=small,labelfont=bf]{caption}

\begin{document}
\title{\name: Batteryless, Thin Sticker-like Flexible Force Sensor}


\author[test]{Agrim Gupta$^*$, Daegue Park$^o$, Shayaun Bashar$^*$, Cedric Girerd$^o$, Tania Morimoto$^o$, Dinesh Bharadia$^*$}
\affiliation{$^*$: Electrical \& Computer Engineering, $^o$: Mechanical and Aerospace Engineering, UC San Diego}
\email{(agg003,d8park,sbashar,cgirerd,tkmorimoto,dineshb)@ucsd.edu}

\ifimwut
\begin{abstract}

Any two objects in contact with each other exert a force that could be simply due to gravity or mechanical contact, such as a robotic arm gripping an object or even the contact between two bones at our knee joints. The ability to naturally measure and monitor these contact forces allows a plethora of applications from warehouse management (detect faulty packages based on weights) to robotics (making a robotic arms' grip as sensitive as human skin) and healthcare (knee-implants). It is challenging to design a ubiquitous force sensor that can be used naturally for all these applications. First, the sensor should be small enough to fit in narrow joints or not make the robotic arm bulky. Next, we don't want to lay long, cumbersome cables to read the force values from the sensor to the monitoring device. And finally, even current wireless technology requires a battery that could be hazardous for in-body applications. We develop WiForceSticker, a wireless, battery-free, sticker-like force sensor that can be ubiquitously deployed on any surface, such as all warehouse packages, robotic arms, and knee joints. WiForceSticker first designs a tiny $4$~mm~$\times$~$2$~mm~$\times$~$0.4$~mm capacitative sensor design equipped with a $10$~mm~$\times$~$10$~mm antenna designed on a flexible PCB substrate. Secondly, it introduces a new mechanism to transduce the force information on ambient RF radiations that can be read by a remotely located reader wirelessly without requiring any battery or active components at the force sensor, by interfacing the sensors with COTS RFID systems. The sensor can detect forces in the range of $0$-$6$~N with sensing accuracy of $<0.5$~N across multiple testing environments and evaluated with over $10,000$ varying force level presses on the sensor.
We also showcase two application case studies with our designed sensors, weighing warehouse packages and sensing forces applied by bone joints.

\end{abstract}



\fi

\maketitle

\begin{figure}
    \centering
    \includegraphics[width=0.95\textwidth]{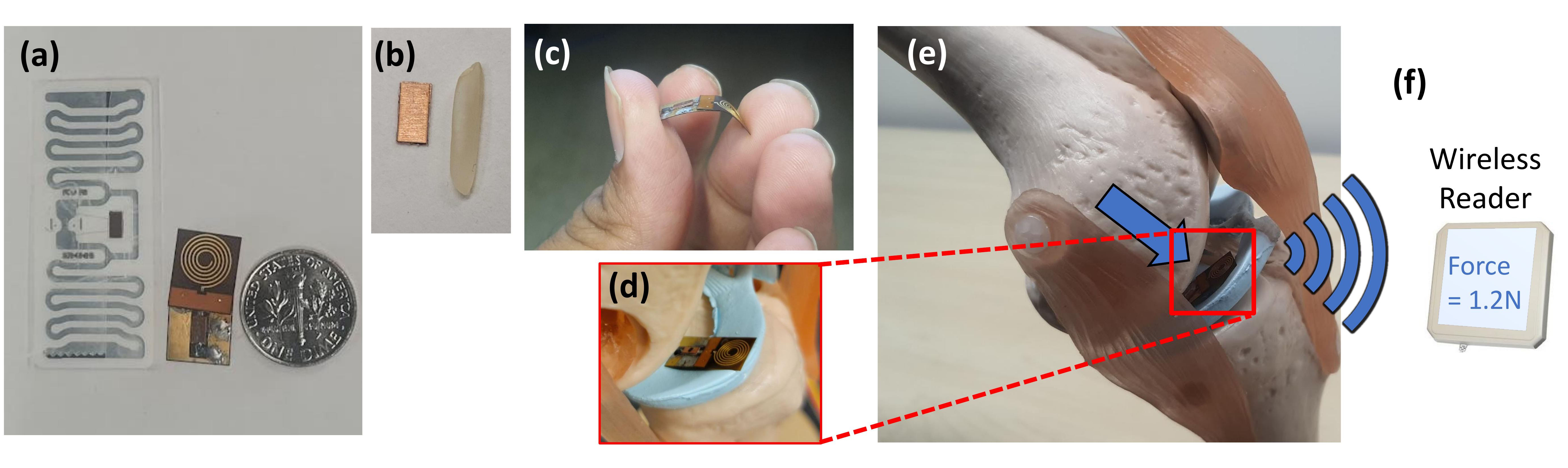}
    \caption{(a) Shows 2 versions of \name, integrated with standard RFID stickers and a small flexible PCB (b) Shows the sensor by itself, compared to a rice grain (c) Demonstrates the flexibility of the sensor PCB designed (d,e,f) Shows how \name can be used for in-vivo implant force sensing read wirelessly via an remotely located reader}
    \label{fig:intro_fig}
\end{figure}
\section{Introduction}
\noindent Force is a ubiquitous phenomenon in our daily environment, with any two objects in contact exerting forces onto each other. 
Therefore, force sensors are a simple and unique way to quantify ubiquitous contact phenomenon and bridge the gap between the digital world and the real environment, with clear use cases in AR/VR and ubiquitous computing~\cite{bai2020stretchable,zhu2020haptic,dangxiao2019haptic}.
Further, these force sensors can enable a myriad of diverse applications, from warehouse scenarios, orthopedic implants to robotics.
For example, we could sense the weight (force) of the packages in a warehouse setting and use the sensed force for the integrity of the packages~\cite{sensor_checking_weights}, since an empty or improperly packed package would apply a lower gravitational force. Similarly, a prosthetic orthopedic implant equipped with a force sensor could measure the contact forces between the knee joints as the patient recovers post-surgery to categorize the implant health (shown in Figure~\ref{fig:intro_fig}{(d-f)}) and how well the body accepts the knee implant~\cite{sensor_check_knee_implants_1,sensor_check_knee_implants_2}. Another application area is robotics, where these force sensors can be used to better control the robots, for example while gripping a sensitive object, the robotic manipulator needs to quantify the forces it is applying~\cite{gripper_force_needed_1_fruit,gripper_force_needed_2_robotic}, and in surgical robots, which have to be careful while interacting with body tissues so that they don't damage the tissues by applying excessive forces~\cite{force_needed_1_surgical_robot}. 

However, these applications pose numerous challenges to force sensors, and hence the applications are not mainstream today. 
For example, consider the use case of sensing forces sustained by the knee implant.
In this scenario, the force sensors need to be thin enough to fit within the narrow gaps between the bones in the knee joint as shown in Figure~\ref{fig:intro_fig}{d}.
Further, it needs to fit snugly to the uneven bone shapes in the knee joint.
Since the implant is located inside the body (in-vivo), it is not possible to carry a wire to the sensor, and thus the sensor needs to be read wirelessly via a remotely located reader outside the body as demonstrated in Figure~\ref{fig:intro_fig}{f}.
To enable wireless operation, we can not assume an active radio near the sensor to transmit the sensor readings. As an active radio would require a battery, and batteries pose a risk of contamination inside the body, as well as the requirement of a battery would not allow for thin sticker-like sensors.
Hence, the ideal force sensor needs to be `thin' and have a `sticker-like' form factor to fit snugly inside the joint and at the same time support `wireless' operation without requiring a `battery'.


To achieve these requirements, a few existing mm-scale sensors typically create voltage fluctuations that need to be sensed locally via amplifiers and ADCs~\cite{dahiya2011towards,ibrahim2018experimental}, prohibiting these sensors' from being energy efficient. Hence, these existing sensors cannot be used in a batteryless sticker-like form factor. Another recent work attempts to bypass the need for amplifiers and ADCs by directly creating signal fluctuations proportional to force applied on a surface onto an incident Radio Frequency (RF) signal. These signal fluctuations cause a shift in the phase of the reflected signal, which is used by a remotely located reader to infer the force reading~\cite{wiforce}. However, the sensor designed is rather quite large (80mm) to be able to create measurable signal phase changes. More specifically, the sensor length (size) is proportional to the changes in the signal phase. Therefore, this makes it extremely challenging to create a tiny sub-mm sized sensor, deformations of which by applied forces create substantial signal fluctuations to be read by a remotely located reader.


In this paper, we present \name, which for the first time demonstrates these wireless force-sensitive stickers (Fig. \ref{fig:intro_fig}{(a-d)}), which can be stuck to everyday objects, as well as flexibly conform to orthopedic implants (Fig. \ref{fig:intro_fig}d,e) and robots, in order to meet these various applications. 
Specifically, \name develops a novel force-sensitive sensor, which is mm-scale, as shown in Figure~\ref{fig:intro_fig}{(a-c)}, is energy efficient, and develops a novel force transduction mechanism that enables measurable changes in the RF signals as the sensor deforms under applied forces. 
Furthermore, \name uniquely interfaces these mm-scale force sensors to both existing commericla and custom designed RFID stickers, which are flexible and allow for wireless interrogation via a remotely located RFID reader.
\name finally develops robust signal processing algorithms atop of the COTS RFID reader to allow for reliable reading of applied forces on the mm-scale sensor in a variety of environments and even in the presence of moving objects in the environment, to achieve the vision of ubiquitous force sensing. 
This method of creating force-sensitive RFID stickers easily meets the motivated applications of knee implant force sensing and ubiquitous weight sensing in warehouses and has the potential to enable many more applications.

The first challenge for creation of such force-sensitive stickers is to design a mm-scale sensor that creates measurable changes in the wireless signals reflecting off from the sensor.
Under the applied forces, the mm-scale force sensor would create even smaller sub-mm deformations in the sensor geometry. These sub-mm deformations needs to be sensed remotely at readers located meters away from the sensor, which are at least 1000 times more distance than the created sub-mm deformation.
A typical approach for sensing these sub-mm deformations is to measure the phase change, which is proportional to change in length. However, such approach wouldn't work at <1 GHz frequency, where wavelengths are larger than 300 mm\footnote{Above 1 GHz signals don't penetrate in-vivo}, which creates sub-1$^o$ phase changes to measure sub-mm deformations, which are almost impossible to measure.
\name solves this challenge by identifying that sub-mm deformations along the sensor thickness instead of sensor length provides an attractive alternate mechanism.
Essentially, change in sensor thickness due to applied force creates capacitance changes, and these sub-mm deformations have to travel through a different medium rather than air, which creates novel effects due to boundary effects of changing materials and hence leads to measurable changes even when deformations are sub-mm. 
\name uses this capacitance change phenomenon to create a new transduction mechanism that translates the load capacitance into phase change at RFID frequencies (around 1 GHz), which essentially enables a measurable $>10^{o}$ phase change, even from the sub-mm deformations, when the sensor is miniaturized to mm-scale.

We provide a mathematical understanding of this new transduction mechanism and show a framework which can be used to design such mm-scale capacitive force sensors which transduces force to RF signal phases.
Said differently, we choose the sensor dimensions, the dielectric layer thickness and material to ensure that the physics of transduction mechanism works out at the chosen frequency of 900 MHz.
Further, we perform multiphysics simulations in order to solve both maxwell's equations for force impedance calculations, as well as solid mechanics equations for force deformation in order to model the transduction phenomenon and confirm the sensor design, as well as the required RF interfacing, to guarantee a measurable wireless channel phase change for the joint communication sensing paradigm proposed by \name. Thus, \name creates a powerful joint communication-sensing paradigm for force sensors that can scale to mm-scale sensors, and would be energy efficient, and modulate the RF signals with measurable phase shifts as a function to applied force.
 




Next, we design an interface for \name sensor to the backscatter technology like RFIDs, which can modulate the force-to-phase shift onto the wireless channel of RFID tags and enable battery-free operation. 
That is, it gives a discernable identity to the signals reflecting from the sensor via the RFID modulation which allows for the RFID reader to decode these reflected signals and compute the phase shifts as well to measure forces wirelessly.
Furthermore, this RFID integration also enables \name to support multiple force sensors concurrently as well, with RFID providing unique identity to each. 
\name also makes a key observation that the \name sensor is designed to be purely capacitive, and thus has purely reactive impedance. 
When a purely reactive impedance is interfaced in parallel to the RFIDs, we obtain a pure phase shifted signal with no reduction in the magnitude of reflected signal.
Since \name sensor only affects phase and not magnitude, it does not disturb the operation of RFID energy harvester and hence does not reduce the RFID reading range.
This interfacing with RFIDs without losing any range allows for robust wireless sensing which does not compromise the reading range, or the sensor signal strength.
Further, \name also creates custom RFID tags on flexible PCB substrates, and we utilize a self designed small 1cm$\times$1cm spiral PCB printed antenna to further miniaturize the form factor of the sensor.
We also show how we can interface the \name sensor with existing commercialized RFID strickers in a standalone manner without requiring a PCB.

Finally, we design novel signal processing algorithms to read the phase changes from the sensor using COTS RFID readers, which would enable immediately usablity of \name. \name develops the force measurement via phase readings reported by the RFID reader. FCC compliant RFID readers hop across the 900-930 MHz RFID band every 200ms, which creates phase discontinuities in the reported phase. However, since \name sensor conveys the force information via change of phases, by taking differential phases over time, we can get rid of the random hopping offsets introduced.
This elimination of random phase offsets allows for collating consistent information across the 50 different RFID channels, which in turn allows improved sensing performance because of channel averaging of phases, giving the capability to read phase changes accurate to 0.5-1 degree, which allows to sense forces with 0.2-0.4 N resolution across the 0-6N sensing range.

To summarize, we present first design for a mm-scale force sensor which can transduce force onto wireless signals at RF frequencies and integrate them with flexible RFIDs to attain the sticker form factor battery-less force sensors. We prototype the sensor based on the optimized design in COMSOL for the knee implant force sensing and ubiquitous weight application, using ecoflex material for the sensor fabrication. We integrate the sensor with popular RFID chipsets, along with multiple different flexible antenna design which can be powered and read wirelessly. The fabricated sensor by itself is 1000x lower volume form factor as compared to recent past work~\cite{wiforce} on wireless force sensors and is the first work to showcase an end-to-end battery-less wireless operation. 
We experimental show that our results obtained are consistent with the proposed transduction mechanism of capacitance to phase changes.  Our key results include: $<$0.5 N sensing accuracy, tested over $>$ 10000 contact forces applied on the sensor, both over the air and over a pork belly setup for in-body sensing. We also test the sensor with people moving around and show that our signal processing algorithm has robustness of phase sensing to the dynamic multi-path as well. Furthermore, our experimental prototype matches closely with FEM multiphysics simulations in COMSOL. We also present in simulations using our framework could be used to develop sensor for different ranges of force, by choosing appropriate polymer.

Using the above sensor design, we also conduct two experimental case studies with our sensor. 
In our first case study, we demonstrate reading the forces applied by knee joints using toy-knee kit, and accurately measuring these force even in-body experiments. Finally, in our second case studies, we put different quantities of objects in a package and read weight from the force stickers attached to the bottom of the package, to identify if the packages are meeting the integrity on number of items in the package. We believe \name could be sensed many more quantities beyond just force, and could be generalized to using even everyday WiFi devices as readers.   



\section{Background and Motivation}\label{sec:background}

In this section, we would present the need for sticker-like wireless readable force sensors which allow scalable deployment of force sensors and enable numerous applications, like ubiquitous weight sensing, and smart in-vivo implants. 
Next, before moving to how we create these force stickers, we show why existing techniques fail to create these stickers and how \name fits in context to past work on such stickable sensor designs.

\subsection{Need for Batteryless and sticker-like force sensors}
\begin{figure}[t]
    \centering
    \includegraphics[width=\textwidth]{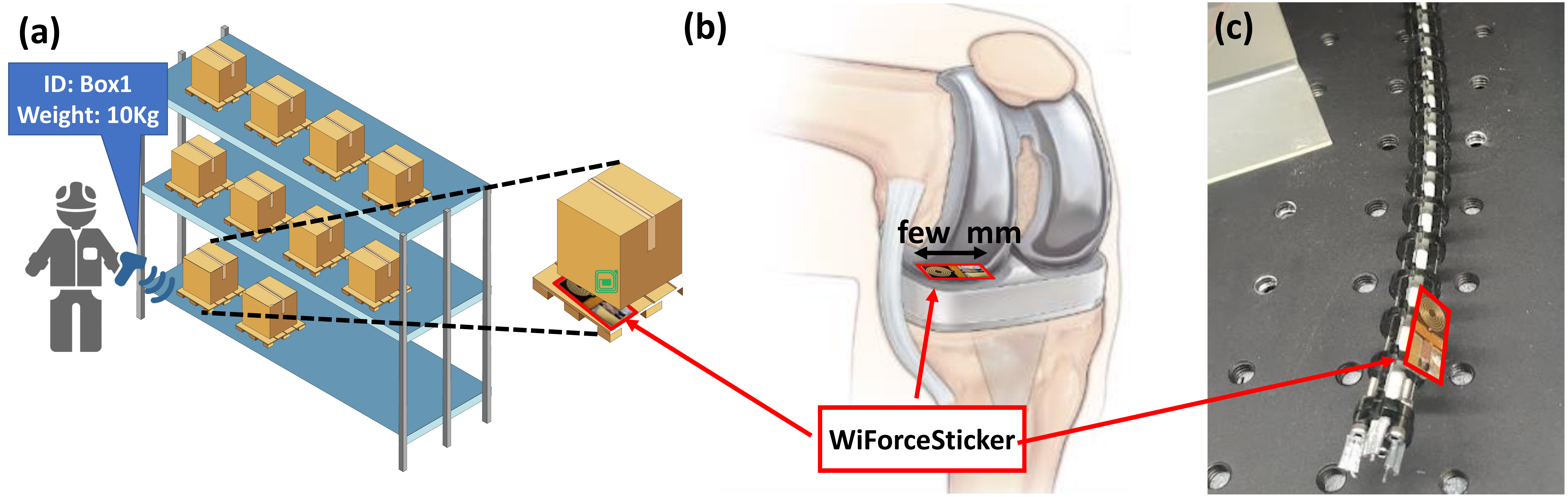}
    \caption{(a) Shows how \name can be stuck on bottom of packages to measure the weight of contents inside for integrity checks (b) Shows how \name can be used for in-vivo implant force sensing and (c) Shows a possible integration of \name sensor with tendon driven robots}
    \label{fig:bgm_fig}
\end{figure}
Any two objects in contact exert forces onto each other, and thus in order to measure these contact forces it is highly imperative that the force sensors be thin form factor to fit in between the two objects without disturbing the contact process by much.
Further, these objects need not have flat rectangular contacts, for example, our bone joints have bones in contact with each other via extremely irregular shapes.
Now, in order to confirm well to various such objects shape, these thin sensors need to be flexible and fit snugly on these objects, akin to small form factor thin stickers.
Hence, an ideal force sensor to sense contact forces between any two ubiquitous object is a small, flexible and thin form-factor sticker-like sensor, and these force sensitive stickers would open up a plethora of applications.

In a warehouse setting, these force stickers can be stuck to bottom of packages and this will allow a portable weighing machine in form of these sticker readers to ubiquitously read the weights of the content similar to how bar-codes readers work as shown in Figure~\ref{fig:bgm_fig}(a).
This feature can then be used to determine empty or incorrectly packed packages.
A naive solution with existing force sensors would be to attach small weighing scales below each package. 
However this is clearly not scalable, given the form factor of such sensors which will increase the packet size, cost of putting a sensor on each package and manual reading of these multiple weighing scales.
On the other hand, a sticker like battery-free force sensor, which can be read as easily as a barcode, would be a scalable solution as it will solve all the challenges: thin paper-like form factor, negligible cost, and easy automated reading of weight via a scanner like a device.

Due to the thin form factor and batteryless feature, these force stickers would be apt candidates to enable smart force-sensitive implants. 
Today, the field of medical orthopedic implants is already looking to engineering fields for innovation, with personalized 3-D printed implants making a lot of headway~\cite{impl1,impl2,impl3}.
The implants can be made much smarter by adding these force stickers on top of these 3-d printed structures; since these are highly flexible, the sticker would conform well to the implant's irregular personalized shape and be able to measure the bone joint forces as shown in Figure~\ref{fig:bgm_fig}(b).
There are already research studies motivating the use of force sensors in implants to enable implant health monitoring and sense how well the implant fits within the bone joint~\cite{safaei2018force,s140815009,knee_implant}.
However, these studies were done in laboratory settings with wired force sensors which can not work in-vivo.
These sticker-like force sensors being batteryless would finally enable these in-vivo applications in addition to the ubiquitous weight sensing application discussed prior.

Another application which these batteryless force sensitive stickers can enable involves sensing forces applied by robots to improve their control and operation~\cite{yip2016model,billard2019trends, deng2020grasping} as shown in Figure~\ref{fig:bgm_fig}(c).
Wired force sensors could hamper the robot's movements and make the robot less flexible.
The problem gets exacerbated with emerging robot designs like soft robots~\cite{burgner2015continuum}, which have a highly flexible and small form factor since they are often designed for use in surgery \& search and rescue, where they must navigate through highly constrained environments.
Hence, the ideal force sensors for these robotic applications must also be small and flexible, and the presence of batteries or wired links makes these requirements difficult to meet.
Such requirements again motivate the force stickers, which can be stuck to various parts of robot without disturbing the robot motion by much and give the required force feedback for efficient robot control.

\subsection{Can existing sensors be made wireless and sticker-like form factor for these applications?}
The existing force sensors are typically MEMS-based discrete sensors that use a variety of transduction mechanisms, like the force-induced change of resistance/capacitance (resistive/capacitive sensors)~\cite{capacitivesurvey,capacitivetina,resistive1,resistive2} and piezo current generated due to applied force (piezoelectric sensors)~\cite{piezo3,piezo2,piezocatheter}. 
The typical paradigm is to digitize the transduction effect via Analog to Digital Converters (ADCs)~\cite{dahiya2011towards,ibrahim2018experimental} or Capacitance to Digital Converters (CDCs)~\cite{capacitivetina,kim2015force}, which measure the change in current/resistance and capacitance, respectively.
Now to communicate wirelessly, these digitized values are typically transmitted via digital communication blocks like BLE/Wi-Fi, etc.
This interfacing requires a battery which ends up violating the form factor requirements for sticker-like wireless sensors.
As a consequence, the applications for such force sensors have been limited to those that have the resources for a battery, for example, weighing scales with load cells~\cite{weigh_scale}, force-sensitive resistors beneath our trackpads~\cite{sensel} and capacitive force sensors in latest AirPods for haptic-based volume control~\cite{airpods}. 

Now to make these sensors sticker-like form factor and batteryless, an attractive approach is to interface the sensors to RFIDs.
RFIDs already come in sticker like form-factors and communicate to a RFID reader wihtout any battery by relying on energy harvested via RF signals.
An approach tried by past work ~\cite{omidRFID} is to directly connecting COTS sensors to RFID by cutting a small part of the RFID antenna and replacing it with the sensors in series.
Although~\cite{omidRFID} evaluates this strategy for Temperature/Light sensors, it is still possible in the purview of the suggested RFID hacking to put a small COTS force sensor using similar cut and interface in series technique.
This is because the COTS sensors lead to change in sensor impedance due to a change in Temperature/Light, similar to how force sensors change impedance under forces. 
This change in impedance across the cut parts of antenna creates fluctuations in the wake-up thresholds of the energy harvester, and this fluctuation is used as a metric for wireless sensing.
However, this technique hampers the RFID read range in the process as it disturbs the energy harvester, which impacts the quality of the reflected signal and hence the sensor resolution.
In fact, this particular past work~\cite{omidRFID} poses the sensor integration with RFIDs as a challenge paper, so as to how to integrate the sensors without losing range and resolution in the process.

In this paper, we essentially give a solution to the so-posed challenge by designing a new miniaturized force sensor that works as a capacitative sensor. Our key insight is that when we interface this force-sensitive capacitor in parallel to the RFID, it creates pure phase shifts with minimal change in amplitude.
Hence, this does not hamper the amplitude of signal reflections and thus preserves the range of the RFIDs and allows the sensing to go ahead at the highest possible resolution.
Due to this successful RFID integration, we achieved the first force sensors in this sticker-like form factor and demonstrated various applications.


\section{Design}\label{sec:design}

\name presents the first sticker-like, wireless, and batteryless (mm-scale) force sensor.
In this section, we delineate various facets behind \name.
First, we will present the joint communication sensing transduction mechanism, which allows for mm-scale batteryless \name sensor design.
Then, we present the sensor design, which is just 0.4mm thick, allowing for sticker-like form factor, and motivate its working principle via Multiphysics simulations.
Going ahead, we show how the designed force sensor can be interfaced with RFID ICs and existing RFID stickers, which modulate the sensor signal.
Finally, we present the details on how an externally located RFID reader isolates the modulated sensor signal and reads the phase shifts to measure the forces wirelessly from the sensor.


\subsection{A joint communication sensing paradigm for mm-scale force sensors}

\begin{figure}[t]
    \centering
    \includegraphics[width=0.8\textwidth]{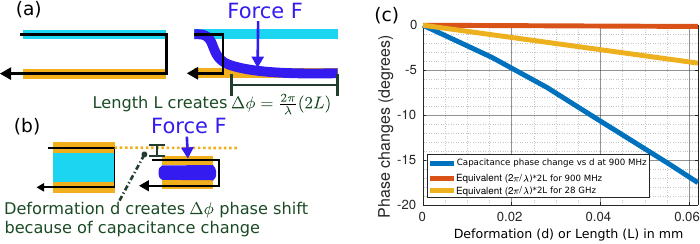}
    \caption{(a) Shows a visual comparison between \name's capacitive transduction mechanism and past work's~\cite{wiforce} length based mechanism (b) Shows how the capacitive approach allows for sensor miniaturization with measurable phase changes even when deformations are sub-mm scale}
    \label{fig:design_intro}
\end{figure}

In order to successfully enable a sticker-like form factor wireless force sensor, one key requirement is to have a batteryless operation.
The presence of a battery would occupy substantial space and thickness, and not permit simple force-sensitive stickers.
To enable batteryless operation, a major design choice is to create a joint communication sensing paradigm so that both sensing and wireless communication happen simultaneously by encoding force information directly onto wireless signal fluctuations. 
These fluctuations could be signal amplitude decrease by absorbing certain frequencies, phase shifts in the channel or polarization changes. 
Changes in amplitude/polarization are often not generalizable to different environments as there are other objects in the environment as well, which absorb certain frequencies.
However, the phase shift is a robust parameter that can be read very accurately with modern radios and has been extensively used in the past to support various joint communication sensing paradigms~\cite{li2016paperid,pradhan2020rtsense,wiforce}.

However, designing phase shift based joint communication sensing paradigm is challenging for mm-scale sensors.
For context, a popular method used in past works~\cite{li2016paperid,wiforce} is to use the well-known linear relationship between phase shift, $\phi$ and the distance travelled by the wireless signals, $d$ having certain wavelength $\lambda$, where $\phi(d) = \frac{2\pi}{\lambda}d$.
The sensor in~\cite{wiforce} was $80$ mm long and worked by having a flexible transmission line separated by an air gap, which allows the line to bend in response to the applied force and create deformations along the length (Fig. \ref{fig:design_intro} (a)).
Now, hypothetically, even if the sensor is miniaturized successfully to say $1$mm long sensing strip, and assuming that at $0$ force applied the signals just reflect back near the end of sensor, whereas when highest force is applied, the maximum possible $1$mm deformation is created such that signals reflect back near the start of the sensor, the maximum phase shift even in this best possible scenario would be $\sim 2^{o}$ at 900 MHz frequencies, having $33$cm wavelength.
In actuality, the bending will not be as effective as the entire sensor length, with the $80$ mm sensor prototype showing only $1/8$-th effective bending of $10$mm. Thus a hypothetical $1$ mm long sensor based on past work would lead to even lower sub-mm deformations and lesser than $1^{o}$ phase shifts.
A naive solution to this would be to scale to higher frequencies like 28 GHz, which have mm-scale wavelength would give $>10^{o}$ phase shifts which can be measured. 
However, scaling to higher frequencies will make the sensor impractical for in-vivo applications, as well as make the sensor hardware extremely difficult to design and integrate.
Further, the presence of an air gap to allow for bending also limits the application of this sensor, as it can not be fabricated in a sticker-like form factor even if miniaturized successfully at high frequencies.


The insight \name has to address these problems with length-based force to phase transduction is to replace the air gap between two conductive surfaces with a soft deformable polymer, which compresses under the action of force (Fig. \ref{fig:design_intro}b).
This deformation creates changes in the capacitance of the sensor as the two conductors get closer to each other because of polymer deformation, which results in phase changes.
Fundamentally, when we go to this new transduction mechanism of capacitance to phase, we now use phase relationships to deformations along with the sensor thickness instead of sensor length.
Consequently, the transduction mechanism does not depend much on the sensor length and instead depends on sensor thickness, thus allowing the sensor to be made with very small form factors.
In addition, phase change along the thickness creates measurable phase changes even when deformations go to sub-mm levels (lesser than a typical thickness of paper strip, which is around 0.1mm), as shown in (Fig. \ref{fig:design_intro} (c)) so that the sensors are not very thick and can be made in a sticker form factor.

This phenomenon of change in phase due to conducting surfaces coming closer to each other due to polymer deformation, which leads to a change in capacitance, draws a parallel to varactor-based RF phase shifters popular in antenna arrays. 
Varactors are basically small tunable capacitors and produce a programmable phase shift depending on the capacitance value~\cite{varactor1,skyworks_varactor}.
Thus, using a similar mechanism, \name used the relationship between capacitance and signal phase to create a joint communication sensing paradigm for force sensing, which scales well to mm-scale sensors without increasing operating frequency.
In the next section, we will explain more about this phase change phenomenon, how to model it mathematically and use this modeling to design the mm-scale \name.

\subsection{How to design mm-scale capacitive sensors to give measurable phase change?}

\begin{figure}[t]
    \centering
    \includegraphics[width=0.8\textwidth]{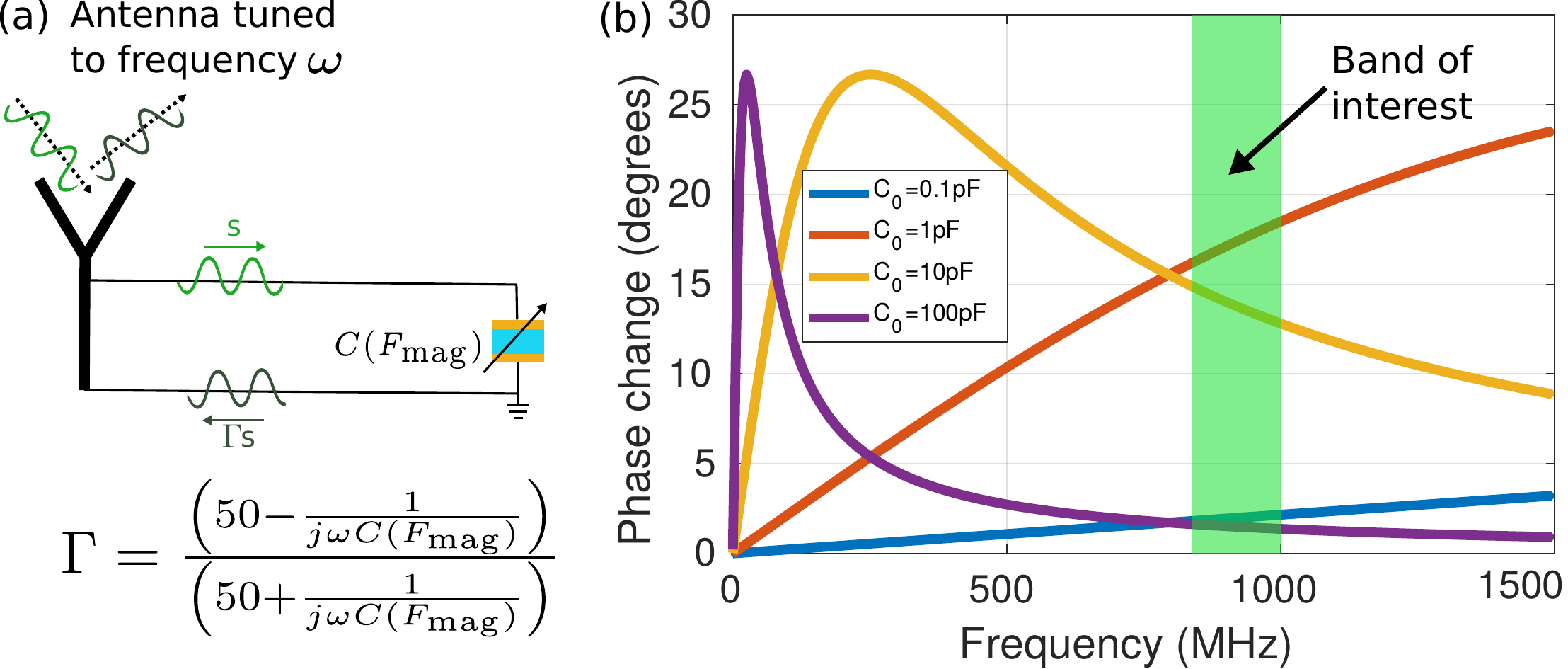}
    \caption{(a) Reflection coefficient equation (b) How different values of $C_0$ affect $\Delta \phi$}
    \label{fig:delphi}
\end{figure}

We start by representing the sensor capacitance as a function of force applied on it, $C(F_{\text{mag}})$.
Say if we have an antenna receiving signals $s$ at RF frequency $\omega$, and a transmission line 50 $\Omega$ matched to $\omega$, and we terminate this with the capacitive sensor (basically like a pure reactive load), as shown in Fig. \ref{fig:delphi} (a).


The reflected signals from the capacitor are multiplied by the reflection coefficient $\Gamma$ and then backscattered by the antenna. Notice that due to the purely capacitive nature of the sensor $\Gamma$ has the form of $\frac{a-bj}{a+bj}$ 
and hence can be rewritten in polar form as $\frac{\sqrt{a^2+b^2}e^{-j\tan^{-1}(b/a)}}{\sqrt{a^2+b^2}e^{j\tan^{-1}(b/a)}}$.
Thus, $\Gamma$ has unit magnitude and a phase term $\phi(F_{\text{mag}})$ given by $2 tan^{-1}(b/a)$, which evaluates to

\begin{equation}
\phi(F_{\text{mag}}) = 2 tan^{-1}(1/50\omega C(F_{\text{mag}}))
\label{eq:phase}
\end{equation}

Hence, capacitance change as a function of applied force leads to a change in the reflected signal phase when interrogated at an RF frequency $\omega$, as described by the equation \eqref{eq:phase}.

Typically, the capacitors are used at low frequencies (a few kHz) and capacitance is sensed by measuring the charge accumulation via a CDC (Capacitance to Digital Converter). A general trend is that the sensor sensitivity and performance improve as the capacitance increases.
This is because the charge accumulation, $Q_{\text{accumulated}}=C(F_{\text{mag}})V_{\text{applied}}$ is directly related to capacitance, and thus the charge accumulation due to higher capacitance is more readily measured.
Hence, it is favorable to design sensors with higher capacitances, and a common design theme is to have higher dielectric constant polymers or higher area conductors to facilitate increased capacitance~\cite{jang2016enhanced,capacitivesurvey,capacitivetina}.
However, unlike the low-frequency counterparts, RF force capacitors are to be read via phase changes, which is not a simple linear equation like charge accumulation, and hence these RF force capacitors have to be designed carefully. 
The quantity of interest while designing these RF force capacitors is the total phase change ($\Delta \phi$) due to the maximum permissible force for the sensor $F_{\text{max}}$, and $C_0$ is the nominal capacitance of the sensor at $0$ N force:
\begin{equation}
\Delta \phi = 2 tan^{-1}(1/50\omega C(F_{\text{max}}))-2 tan^{-1}(1/50\omega C_0))
\label{eqn:delphi}
\end{equation}
To get the best sensitiveness, we need to design the capacitance such that $\Delta \phi$ is maximized for the operating frequency $\omega$.
Observe that $\Delta \phi$ has a non-linear arctan function which does not make the sensor design as straightforward as maximizing capacitance for maximizing charge accumulation.
To understand the design choice behind RF Force capacitor, we plot $\Delta \phi$  in Fig. \ref{fig:delphi}(b) for an empirical relationship between $C(F_{\text{max}})=1.75*C_0$, (i.e. $\frac{\Delta C}{C_0}$=0.75, a common operating point of the capacitive sensors \cite{capacitivetina,capacitivesurvey}), and different ranges of $C_0$ values.

We observe from Fig. \ref{fig:delphi}(b) that for $C_0 = 0.1$ pF, $100$ pF the total phase change achievable at the 900 MHz
frequency of interest would be $<5^{o}$ which is not a good operating point.
Intuitively what is happening is that \eqref{eqn:delphi} has the arctan function, which saturates as the input gets closer to both $0$ and $\infty$, and hence at $\omega \to 0$, the total phase changes are negligible as the input is driven to $\infty$ and at high $C_0 \omega \to \infty$ the input goes to $0$ and that also gives lesser phase changes.
Hence, the RF force capacitor can not just maximize the capacitance blindly and has to be designed at reasonable nominal capacitances.
We get reasonable phase changes $>15^{o}$ for $C_0 = 1,10$ pF so the ideal operating point of the sensor is to get the nominal capacitance within this range.

\begin{figure}[t]
    \centering
    \includegraphics[width=0.9\textwidth]{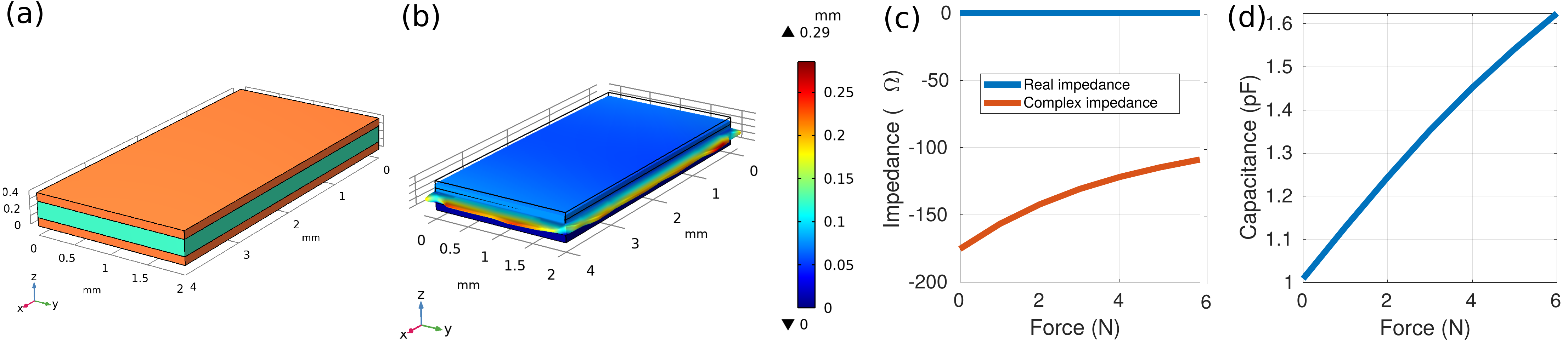}
    \caption{(a) Sensor mechanical composition (b) Deformation under action of force (c) Sensor impedance (real, complex) vs Force (d) Capacitance versus Force simulations}
    \label{fig:sensor_def}
\end{figure}
Hence, to get sufficient phase changes at 900 MHz RF frequencies, we need to design a capacitor with nominal capacitance $C_0$ being of the order of 1-10 pF.
Nominal capacitances can be approximated to be $C=\frac{A \epsilon_r}{d}$ as under no force the sensor is essentially like a parallel plate capacitor, with a Ecoflex 00-30 polymer dielectric layer sandwiched between two copper layers (Fig. \ref{fig:sensor_def}(a)).
The choice of dimensions of the sensor gives a rough initial capacitance of $1 pF$, with $C=\frac{A \epsilon_r \epsilon_{\text o}}{d} = \frac{8*10^{-6}*2.8*8.85*10^{-12}}{0.2*10^-3} = 0.99 pF$, with a $2$mm$\times4$mm = $8$mm$^2$ area sensor with $0.2$mm thick dielectric layer, and the dielectric constant = 2.8 for the chosen Ecoflex 00-30 polymer~\cite{ecoflex_relative_permittivity}.
However, in order to compute the exact $\Delta \phi$, we also need to compute $C(F_{\text{max}})$, and this requires knowing the displacement caused by force, as well as the capacitance of the deformed sensor, which has a squished polymer layer (Fig. \ref{fig:sensor_def}(b)).
In order to do so, we need to solve both the structural mechanics equations to compute the polymer deformation and then use the deformed geometry for maxwell equations to compute the effective capacitance of the sensor and hence obtain the $\Delta \phi$ and the $C(F_{\text{max}})$ metric to finish the sensor modeling part.

Since the polymer deformation has a non-linear stress-strain relationship, and the maxwells equations are not straightforward to be solved in an analytic closed-form approach, we utilize FEM simulation framework, COMSOL Multiphysics 6.0~\cite{comsol}, to simulate the sensor capacitive effect. 
The software takes the sensor geometry as an input and meshes the geometry to form small elements where the differential equations can be solved numerically, and the final results are then collated across each mesh element's solution.
We use COMSOL Structural Mechanics module for the non-linear elastic modeling of the polymer layer and 
via COMSOL AC/DC module, we can excite the sensor with an AC voltage source and compute the obtained current as affected.
This computation can be done with the sensor being under the effect of various levels of force applied to the sensor.
Then, by simply taking voltage to current ratio we can obtain the sensor impedance as a function of force, $Z(F_{\text{mag}})=V(F_{\text{mag}})/I(F_{\text{mag}})$.
Hence, this allows the computation of sensor impedance versus force readings (Fig. \ref{fig:sensor_def}c). 
Since the resistive component of the impedance is almost 0, and the reactive component is negative, it shows that the sensor is almost purely capacitive.
We obtain the capacitance to force curve as shown in Fig. \ref{fig:sensor_def}d, with the capacitance going from 1 to 1.65 pF as the force on the sensor increases from 0 to 6 N.
This meets our target of sensor design with the nominal capacitance of about 1 pF and the ratio of capacitances $C(F_{max})/C_0 = 1.65$, which is reasonably close to the empirical value of 1.75 we had started our calculations with. 

\subsection{Interfacing \name sensor with RFID for wireless operation}

\begin{figure}[t]
    \centering
    \includegraphics[width=0.5\textwidth]{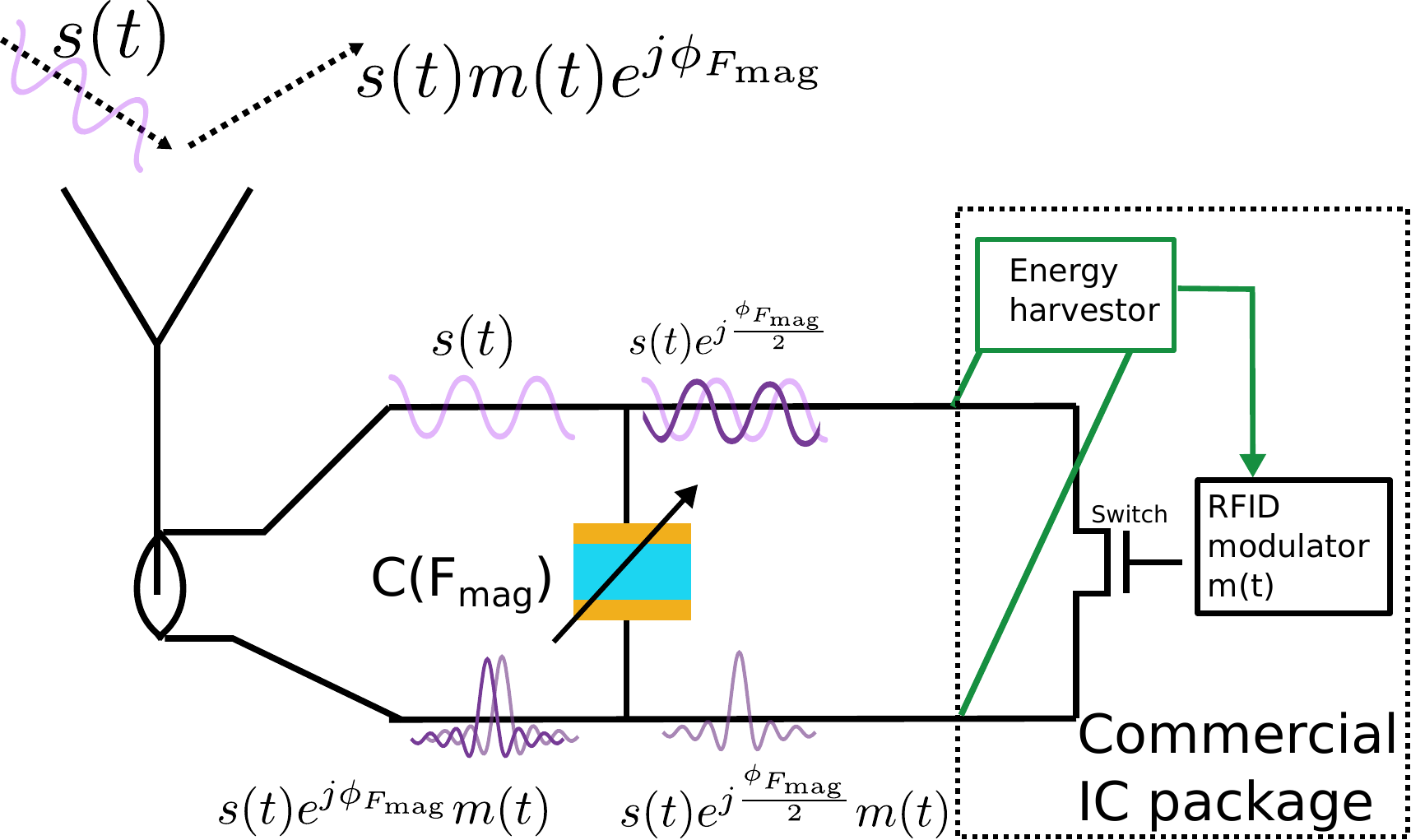}
    \caption{Interfacing the sensor with RFIDs, by inserting the sensor in between the antenna and RFID IC connected in parallel to both}
    \label{fig:sens_modulation_fin}
\end{figure}

\begin{figure}[t]
    \centering
    \includegraphics[width=\textwidth]{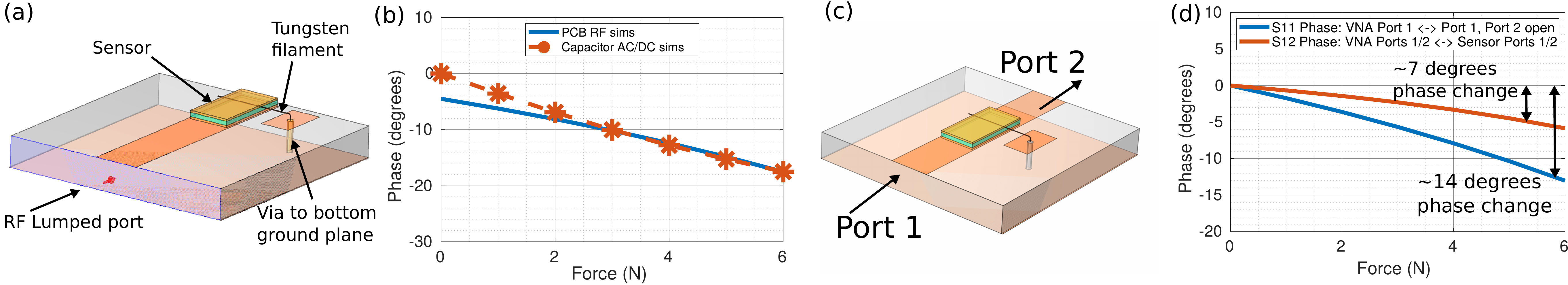}
    \caption{(a) PCB sims with RF module (b) Phase change from RF simulations (blue) are well modelled by the extrapolations from AC/DC simulations (red) (c) Simulating sensor in thru mode (d) Phase doubling effect in simulations}
    \label{fig:sensor_rf_sims}
\end{figure}

For the wireless operation of the sensor, it is not enough to just create phase shifts in the reflected signal.
When a reader situated remotely to the sensor tries to read the reflected signal phases from the sensor, it also needs to isolate the sensor reflections from the myriad of environmental reflections. 
Hence, it is required for the sensor phase-shifted signals also to be modulated before they are reflected back.
To achieve the required modulation, we need to place the sensor in between the antenna and the RFID IC, as shown in Fig. \ref{fig:sens_modulation_fin}. 
However, till now, we have just shown how the sensor is fabricated and mathematically calculated the resultant phase shifts from the changing capacitance of the sensor.
Proceeding ahead, we will show how the sensor is actually interfaced with these RF components such that signals actually pass thru and reflect off from the sensor.



To achieve the sensor excitation via RF frequencies, we simulate the sensor attached to a PCB with a microstrip line matched to 900 MHz frequency using the COMSOL RF module. 
Said simply, this PCB implements a transmission line which has the capacitive sensors as the termination load, in a way it practically implements our capacitive sensing motivation Fig. \ref{fig:delphi}(a).
The PCB consists of two layers; on the top, there is a signal trace which is $2$~mm wide, similar to the sensor, and the bottom layer is the ground plane.
The bottom copper plate of the sensor is directly soldered onto the signal trace.
Now, in order to have the sensor and PCB ground common, we use a $20\mu m$ tungsten wire filament to connect the top copper plate with a small ground pad on the top layer using a via (Fig. \ref{fig:sensor_rf_sims} a). This simulation using the COMSOL RF module can then be linked to the COMSOL Structural Mechanics module to compute the deformations with the updated sensor having a tungsten filament and then calculate the phase from the scattering parameters.
However, this time COMSOL FEM solver solves both the structural mechanics equation and maxwell's equation concerned with the RF signal propagation instead of low frequency AC modeling.
The PCB simulations utilize a single lumped port (Port 1) to excite the transmission line, and the sensor phases are hence encoded onto the S11 measurements in the scattering parameters.

We confirm that the S11 phase changes (about 20$^o$ from RF COMSOL simulations) from the RF module simulations are consistent with the earlier simulations with AC/DC module (Fig. \ref{fig:sensor_rf_sims} b).
Note that the AC/DC module just give capacitance measurements which are mathematically converted to phase via \eqref{eq:phase}, whereas the RF module simulation is a faithful end-to-end FEM simulation which confirms that the sensor can work in a practical form factor. 
The small discrepancies can be attributed to two facts. First, the AC/DC module computes capacitance at small AC frequencies (a few kHz), which may not be purely accurate at 900 MHz. Further, the addition of small tungsten filament may also cause some additional minor discrepancies.
In Section \ref{sec:implementation} we fabricate this PCB and confirm that the RF module simulations are consistent with the actual implementation.
However, in order to add the modulation block, it is required that the sensor also passes the signals through instead of just reflecting them back.
This is required since the existing backscatter modulation blocks, like the commercial ICs available, which implement both energy harvesting units as well as modulation, are designed to be interfaced with an antenna and just reflect back the modulated signal.
To show how \name works in thru mode, we move the sensor in the middle of the PCB simulations (Fig. \ref{fig:sensor_rf_sims} c) compared to towards the end, as shown previously (Fig. \ref{fig:sensor_rf_sims} a). This allows us to define two lumped ports (Port 1, Port 2), one on each end of the PCB, and by observing the S12 phase, we can verify if the sensor can work in a thru mode. 
For this simulation, we obtain the S12 phase changes and observe that the phase changes are about half the magnitude we had before, that is 10$^{o}$ for 0-6 N forces as compared to 20$^{o}$ before. 
Now, when we keep the other port as reflective and estimate the S11 phase, we get almost the same phase shifts as before (around 20$^{o}$).
This thru-mode simulation finally leads to a complete understanding of the sensor's RF level interfacing.
In Section \ref{sec:implementation}, we fabricate the thru PCB on FR4 substrate and interface the sensor with the tungsten filament to also observe this phase doubling effect with actual hardware as well.


Hence, intuitively what happens is that when a signal $s(t)$ travel through the sensor, it first gets a certain value of phase shift and the sensor passes the signal through $s(t)e^{j\phi(F_{\text{mag}})/2}$. Then, when the signals get modulated $s(t)e^{j\phi(F_{\text{mag}})/2}m(t)$, reflect and pass through the sensor again, the total phase shift doubles up $s(t)e^{j\phi(F_{\text{mag}})}m(t)$ and reaches back to the reflective levels as described in Eq.~\eqref{eq:phase}, as visually illustrated in Fig. \ref{fig:sens_modulation_fin}.
Thus, the reflected signal has both the modulation component from the RFID IC as well as a phase shift. 
The RFID reader can use the modulation to isolate the sensor reflections from the environmental reflections and then estimate these phase shifts from channel estimation done on the isolated signal.

\subsection{Putting it all together: Reading forces (phase shifts) from the sensor via COTS RFID readers}



\begin{figure}[t]
    \centering
    \includegraphics[width=\textwidth]{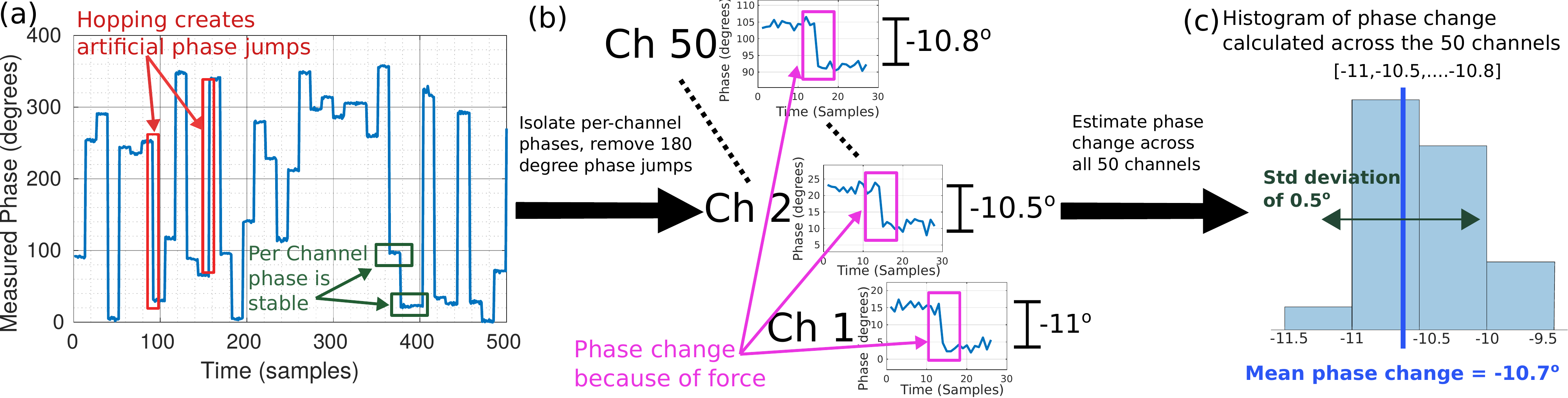}
    \caption{Computing phase change resulting from force applied on the sensor: (a) Shows raw phases from RFID reader (b) Shows per channel phases, by taking differential phase per channel the fixed hopping phases are suppressed and (c) Finally shows how phase jumps are consistent across all the channels}
    \label{fig:sec34}
\end{figure}

To tie up \name's design, the only component left to describe is the wireless reader.
As explained towards the end of the previous subsection, when \name sensor is interfaced with RFID, it reflects back a modulated and phase-shifted signal.
This reflected signal can be decoded via a COTS RFID reader, which is able to sense the phase shifts from the sensor and map it back to applied forces.
The COTS RFID reader achieves this by identifying the signals coming from the sensor via the EPC ID number of the RFID IC integrated with the sensor.
Then, the RFID reader can observe the channel estimates of the particular tag given by the EPC id and calculate sensor phase jumps to estimate applied forces. 

However, in order to estimate these phase jumps with COTS readers, a challenge is the hopping nature of these COTS readers~\cite{tagyro,li2016paperid}. 
As per FCC guidelines, the readers need to change their frequencies after every 200 milliseconds to avoid interference with other readers in the environment. 
The COTS readers introduce a random phase offset due to hopping, as the PLL locks to a different frequency, and this inadvertently shows up as phase jumps in the phase calculations.
However, using the Low Level Reader Protocol (LLRP) for RFID readers~\cite{llrp}, we can isolate the per channel phases since the COTS readers give a channel index for each phase readings, and the phases on a particular channel remain stable across time and do not show these jumps~\cite{tagyro}.
Hence, this allows \name to compute differential phase jumps per channel, which computes the phase shift value caused by force per channel.
Using differential phase automatically offsets the fixed phase jumps across channels and gives a consistent phase jump measured across multiple RFID channels.
In addition to hopping-based phase jumps, the Impinj RFID reader used has phase readings which show 180$^{o}$ shifts and we handle this in our implementation by detecting the phase jumps $>170^{o}$ and removing this 180$^{o}$ appropriately.
Since the force information is encoded in phase jumps which does not exceed jump magnitude of $20^{o}$, we can always detect higher than $170^{o}$ jumps and attribute those to Impinj reader's flaw than force and hence obtain clean jump free phases per RFID channel from the reader.
Then, we can further average the phase jumps recorded across each of the 50 RFID channels between 900-930 MHz to get clean phase jump readings from the RFID reader. This whole process is visually illustrated in Fig. \ref{fig:sec34}.

A unique insight \name has on this RFID sensing is that such multi-channel RFID phase averaging makes the sensing robust to dynamic environmental multipath since each channel records different phase jumps stemming from the dynamic multipath because of its moving nature. However, the phase jump from the sensor remains consistent across the channels.
Thus, upon doing multi-channel averaging across these 50 distinct channels, the multipath phase jumps get averaged out to near zero, and the sensor phase jump remains at its consistent level.
From our experiments, we have observed that in a static environment, such averaging gives phase measurement accurate to $0.5^{o}$ in static scenarios which leads to a sensor resolution of $0.2N$ (Since 0-6N shows approximately $0-15^{o}$ phase jump and hence resolution would be $\frac{6}{15}*0.5 = 0.2$N.
When there is dynamic movement in the environment, the measurements are roughly accurate to $0.8-1^{o}$ (depending on extent of movement) which makes the resolution slightly higher to $\sim 0.3-0.4N$ which is still at sub-N levels.
Further, $0.3N$ is also the reported resolution of forces felt via our fingertips in HCI research~\cite{antoine2017forceedge}, and hence, good enough for many realistic applications.
Hence, \name calculates the phase jumps at each of the RFID hopping frequencies and then averages them to obtain a very clean phase jump measurement (to about $0.5-1^{o}$ precision) which can then be used to compute multiple force levels within the net $15^{o}$ phase jump given by the sensor, with achievable resolution reaching to sub-N levels of $0.2-0.4$N across various environmental scenarios.









\section{Implementation}\label{sec:implementation}

In this section, we first describe the steps required to fabricate the sensor. Then we show how to interface it with actual RF PCBs in simulation using COMSOL RF simulations, and show the consistency between the simulations and our experimental results on a fabricated prototype. Proceeding ahead, we then show three ways of integrating the sensor with RFIDs: using a standard FR4 PCB, an optimized small form factor flexible PCB on polyimide substrate, and on commercial RFID stickers. To conclude the section, we present details on how we implement the wireless reader and required data processing. 

\subsection{Sensor fabrication}

\begin{figure}[t]
    \centering
    \includegraphics[width=\textwidth]{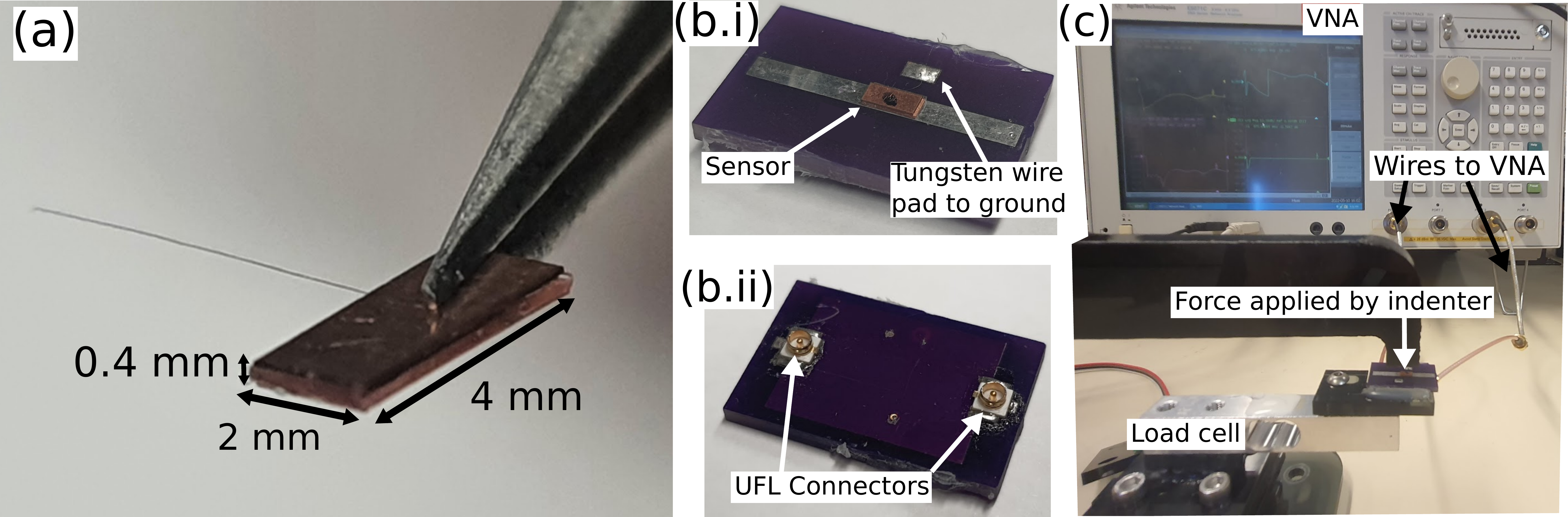}
    \caption{(a) \name sensor after fabrication with the tungsten wire filament (b) Sensor interfaced in parallel to a microstrip line implemented on a PCB, with (i) showing top layer (ii) bottom layer (c) shows the setup to perform wired test at RF frequencies: with forces being applied by an actuated indenter, and ground truth forces measured by the load cell and phases captured by the VNA connected to the PCB via the UFL connectors}
    \label{fig:sensor_vna_setup}
\end{figure}

\name is a capacitive-based force sensor which consists of three layers: a conductive metal layer at the top and the bottom and a dielectric polymer layer in the middle.
When a force is exerted on the conductive layers of the sensor, the dielectric layer deforms to bring the conductive layers closer, increasing the sensor's capacitance.
The Ecoflex-0030 is used for the dielectric layer, whereas copper is used for the conductive layer.
The sensor is $2$~mm in width and $4$~mm in length, and the thicknesses of the copper and Ecoflex 00-30 layers are $0.1$~mm and $0.2$~mm, respectively.
In order to fabricate the sensor, we utilize a blade coating process to get 0.2 mm thick uniform Ecoflex 00-30 layer, which then can be cut into a $2$~mm~$\times$~$4$~mm piece for the sensor.
For the copper layers, we use commercially available $1$~mm thick copper sheet and laser cut them to the required $2$~mm~$\times$~$4$~mm format.

Since the natural adhesion between the copper and Ecoflex 00-30 layers is poor, their contact surfaces are UV/Ozone treated to remove various contaminants on surfaces to improve adhesion.
Then, the bottom copper and Ecoflex 00-30 layer are physically bounded. 
Before bounding the other side of the Ecoflex 00-30 layer to the top copper layer, a commercially available tungsten wire filament~\cite{implementation_tungsten_wire} of diameter of $20$~$\mu$m is placed on the Ecoflex 00-30 layer, for interfacing with the RF components. 
Finally, the top copper layer is placed on the Ecoflex 00-30 layer and the tungsten wire filament.
In addition, this tungsten wire filament has a very small diameter ($100$ times smaller than the thickness of our polymer layer), and thus its impact on the sensor geometry was considered negligible.
The filaments are also flexible and bendable as a consequence of $\mu$m form factor and can be twisted and flexed akin to human hairs.
A zoomed-in image of the fabricated sensor is shown in Fig.~\ref{fig:sensor_vna_setup}(a). For more details on the sensor fabrication process, please refer to~\cite{anon}.

\subsection{RF interfacing of the sensor}
Now, we describe how we obtain wired phase measurements as forces are applied onto the sensor, and confirm the correct behavior of our sensor by comparing the phase-force profile obtained in our COMSOL simulation with our experimental results on our fabricated prototype.
First, we design a microstrip line of $50~\Omega$ impedance at $900$~MHz, that has a width of $2$~mm so that the sensor can be directly soldered on it (see Fig.~\ref{fig:sensor_vna_setup} b.i). The PCB that supports the microstrip line also has a small ground pad for soldering the tungsten wire filament.
The signal traces are taken to the bottom layer using vias, where the line is terminated in two UFL connector pads (Fig.~\ref{fig:sensor_vna_setup} b.ii).
Using the sensor PCB we can interface the sensor via RF cables to a VNA (Vector Network Analyser), a measurement device which mimics a wireless transmitter by exciting the sensor with $900$~MHz signals and observing the reflected and transmitted signals, hence providing wired ground truth phase measurements.
Further, we mount this sensor PCB on a load cell sensor, in order to have ground truth force measurements.
The placement of the UFL connectors on the bottom side of the sensor allows the top surface of the sensor to be free to receive contacts with objects, which he did with an actuated indenter. The RF cables connect the VNA ports to the PCB UFL connectors on the bottom side of the PCB (Fig.~\ref{fig:sensor_vna_setup}c). 

\begin{figure}[t]
    \centering
    \includegraphics[width=0.7\textwidth]{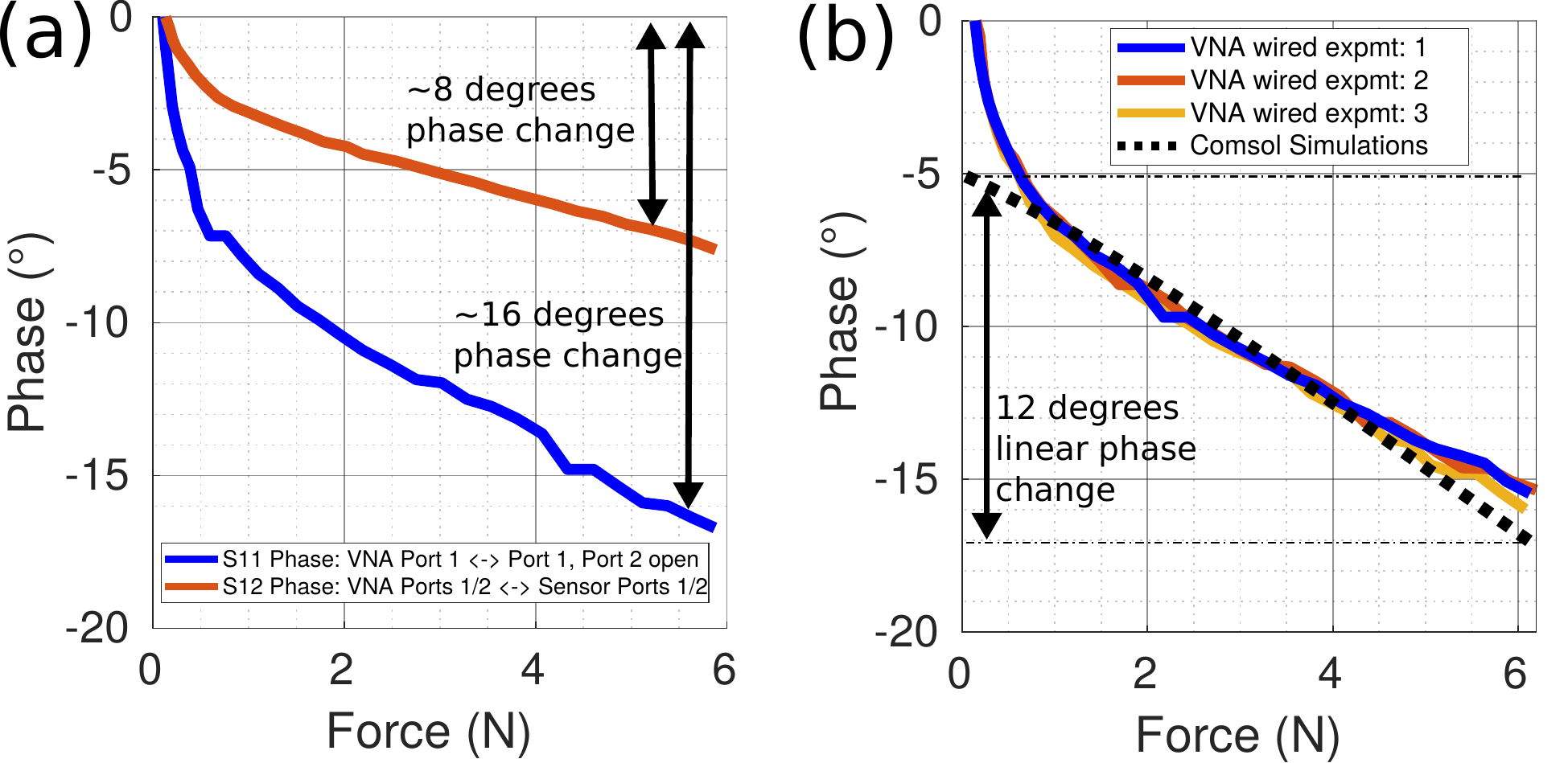}
    \caption{(a) Phase doubling effect due to two sided travel via sensor (b) Consistency with COMSOL RF sims}
    \label{fig:sensor_vna_res_impl}
\end{figure}

In order to compare our physical prototype with our COMSOL simulations, we plot the force to phase characteristic in both reflect (Port 1 connected to VNA and Port 2 opened) and transmit (both Ports 1,2 connected to VNA).
Similar to our intuition as shown in Fig.~\ref{fig:sens_modulation_fin} and COMSOL simulation results as shown in Fig.~\ref{fig:sensor_rf_sims}, we observe that the phase doubling effect happens in our physical prototype (Fig.~\ref{fig:sensor_vna_res_impl}), as from transmit to reflection mode the phases double up from $8^{o}$ phase shift to $16^{o}$. 
Also, we observe consistency with our COMSOL simulations as both simulations as well as hardware results show about 12 degree linear phase change.
However, there is a non-linear part for lower forces ($<1$~N), which is common across capacitive sensors~\cite{sensor_nonlinearity_1,sensor_nonlinearity_2} and can not be modeled accurately via simulations. This initial non-linear part in fact improves the sensor sensitivity in low force ranges.
Also, we confirm that the non-linear response originates from the sensor since even the non-linear part doubles up from transmit to reflect mode, which is only possible if the phase originates from the sensor. 



\subsection{RFID integration with the sensor}
Finally, having described how the sensor is fabricated and how we can evaluate wired phase to force readings to characterize its behavior, we move on to how the sensor force and phase results can be read in a wireless batteryless method by integrating our capacitive force sensor with RFID.

\begin{figure*}[t]
    \centering
    \includegraphics[width=\textwidth]{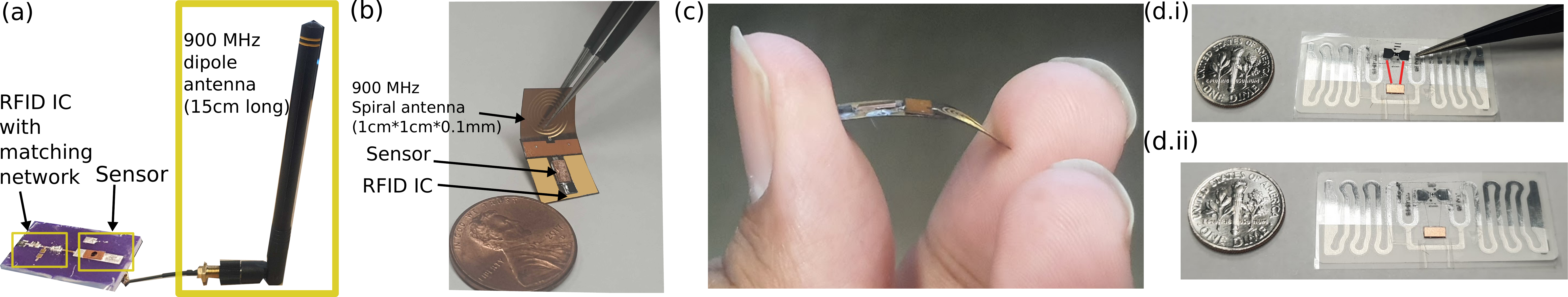}
    \caption{Interfacing the sensor with RFID tags: 
    (a) Shows the rigid PCB with matching network and dipole antenna (b) Shows the optimized spiral antenna flex PCB compared to a dime coin and (c) Shows the flexibility of the spiral antenna PCB and (d) Shows how \name peels the RFID IC from the antenna and attaches the sensor via two tungsten filaments (highlighted in red, d.i) and final standalone RFID sticker (d.ii)
    }
    \label{fig:sensor_rfid_integ}
\end{figure*}

An immediate naive way to achieve this is by continuing further on the PCB and adding an RFID IC pad right after the sensor pad (Fig.~\ref{fig:sensor_rfid_integ}a) and thus implement Fig.~\ref{fig:sens_modulation_fin} on actual hardware. Then, we can connect a simple dipole antenna to the other UFL port and read the sensor wirelessly by looking at the phase of RFID's backscattered signal. For this integration, we use the following RFID IC~\cite{rfidic}. 
However, clearly, integrating with RFIC via a matching network and big dipole antenna is not optimized for space and using such discrete components make the mm-scale sensor bulky.
So, proceeding ahead we miniaturize the antenna as well to fit $1$~cm~$\times$~$1$~cm form factor (Fig.~\ref{fig:sensor_rfid_integ}b).
We design a spiral antenna and use Ansys HFSS to optimize for the spiral dimensions (inner and outer radius), and number of turns in order to get an antenna having around 10\% radiation efficiency at $900$~MHz, which is similar to existing works using $900$~MHz antennas in the $1$~cm~$\times$~$1$~cm form factor~\cite{lee2021msail}.
Further, to avoid matching networks and other connectors, we design sensor interfacing via co-planar waveguides and the spiral antenna connected directly to each other and matched to the RFID tags impedance.
We implement this on a flexible PCB on a polyimide substrate, which is $0.1$~mm thin and can bend and fit various curves (Fig.~\ref{fig:sensor_rfid_integ}c).

In addition to using specialized PCBs for RFID integration, \name sensor can be directly connected to commercial printed RFID stickers (Fig.~\ref{fig:sensor_rfid_integ}d).
These RFID stickers have a peel-able RFID chip which is stuck to the two ports of the RFID antenna.
We can peel the sticker with a tweezer to expose the antenna pads and then interface the sensor via two tungsten filaments, one connected to each of the copper layers directly onto the antenna pads (Fig.~\ref{fig:sensor_rfid_integ}d.i).
Then, we can stick back the RFIC~\ref{fig:sensor_rfid_integ}d.ii) such that the sensor is interfaced again in parallel to both the RFIC and antenna, without actually requiring any additional PCB.

In summary, we have all three options of either interfacing the sensor via a rigid $50$-$\Omega$ matched PCB to an RFID IC, a flexible PCB matched directly to RFID impedance with a PCB printed antenna, which reduces the form factor to serve the in-vivo applications, as well as interfacing the sensor in a PCB-free standalone method to existing commercial RFID stickers, which is slightly larger in area, but similar in thickness to the PCB method, with this sensor interfacing motivating the everyday applications like warehouse sensing in a better way.
Later in evaluations (Section~\ref{sec:results}) we show that all these three integration approaches lead to similar performance for our sensor.

\subsection{Reader implementation}
We use a Impinj Speedway R420 as the reader for our implementation. Since the channel phase measurements contain the information about force, we utilize LLURP protocol to procure phase measurements from the Speedway's FPGA via the SLLURP library in Python 3.6.
The library provides almost real time phase measurements, with callback functions implemented in Python to report fresh estimates as and when provided by the LLURP messaging between the Host PC and the speedway reader.
On our implementation of SLLURP~\cite{sllurp} on a Dell Inspiron laptop with 8 GB RAM, we could get around 80-100 tag readings per second when using Impinj Hybrid Mode 1, which uses Miller 2 encoding to read RFIDs.
For the reader antennas, we use the standard cross-polarized RFID antennas~\cite{alien_ant}.
\section{Evaluation}\label{sec:results}
In this section, we evaluate our \name sensor and present various types of experiments for its evaluation.
We first briefly describe the experimental setup and then present the wireless sensing results, for all three sensor prototypes (rigid PCB, flexible PCB and standalone integration without PCB) shown for RFID interfacing of \name sensor.
Then we show sensor benchmarking studies to showcase the similar error performance of all the three methods of sensor interfacing, durability of the sensor for over $10,000$ force pressing events, and robustness to dynamic multipath in the environment.
Proceeding ahead, we show that \name sensor can be read successfully even when the channel between the RFID reader antenna and sensor is occluded by pork belly, to motivate the use of sensors for in-vivo applications.
Then, we show the range results of \name, by contrasting it with another similar RFID sticker without the sensor to show that integrating sensor with RFID does not tradeoff the RFID reading range.
To conclude, we show \name sensor in action for 2 case studies: sensing knee joint forces and sensing weight of contents in a package to disambiguate number of items inside it.

\subsection{Experimental Setup}
To benchmark the sensor performance, we built the experimental setup shown in Fig.~\ref{fig:exp_setup}.
The setup consists of a linear actuator which applies forces on the sensor via an indenter.
The sensor prototypes (all three different versions, rigid PCB, flexible PCB and commercial RFID) are mounted interchangeably on the load-cell platform such that the applied ground truth forces can be measured by the later, in the same way as for our wired experiments with the VNA.
The RFID antenna is placed directly above the sensor at a distance of $1$~m.

\subsection{Wireless measurement results and sensor benchmarking}

\begin{figure}[t]
    \centering
    \includegraphics[width=0.8\textwidth]{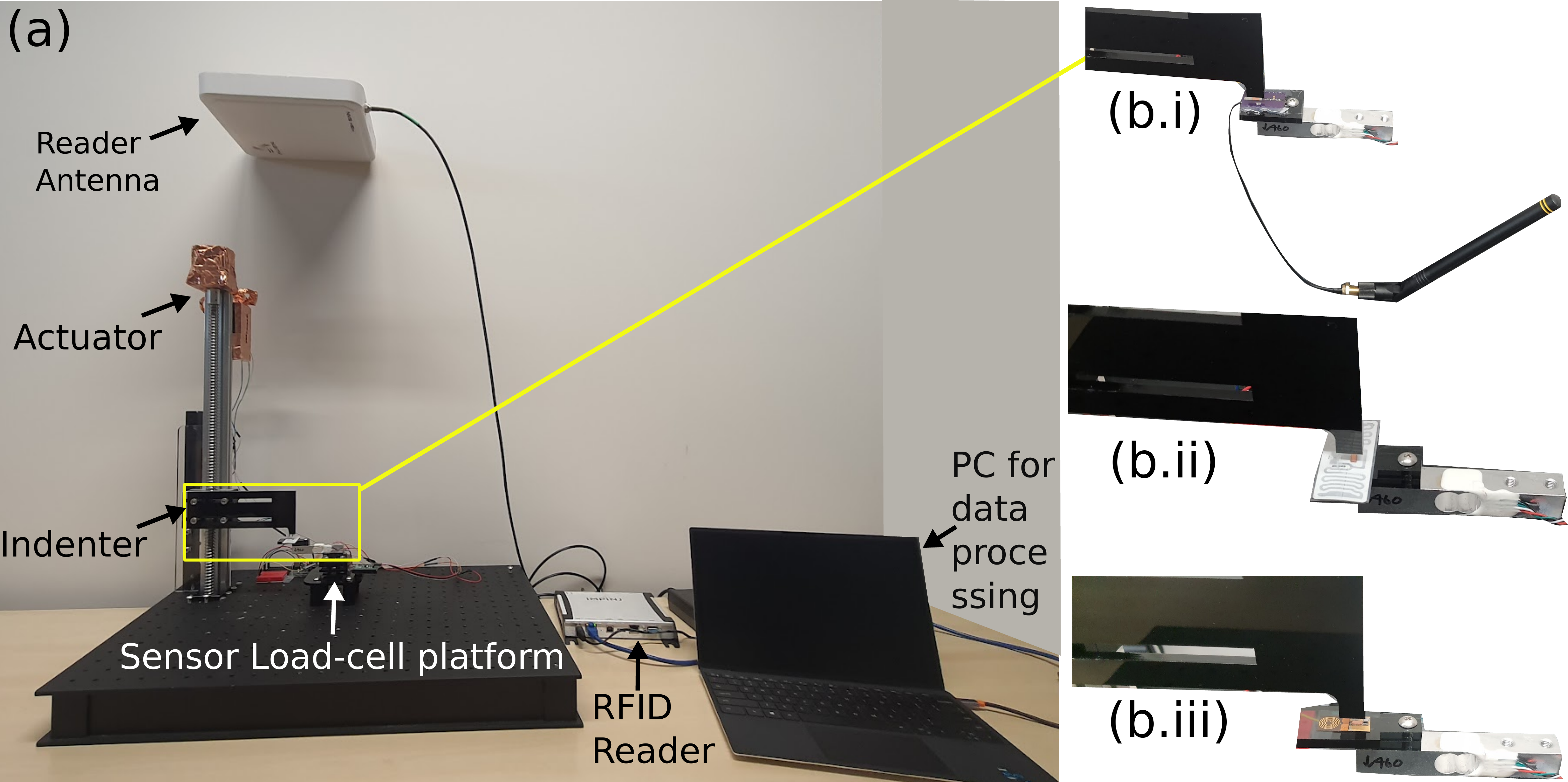}
    \caption{(a) Experimental setup showing the sensor load cell setup, RFID reader and actuator. (b) The sensor under test can be either of the three design prototypes for \name sensor, rigid PCB+dipole antenna (b.i), standalone RFID (b.ii) and flexible PCB with spiral antenna (b.iii)}
    \label{fig:exp_setup}
\end{figure}

\begin{figure}[t]
    \centering
    \includegraphics[width=\textwidth]{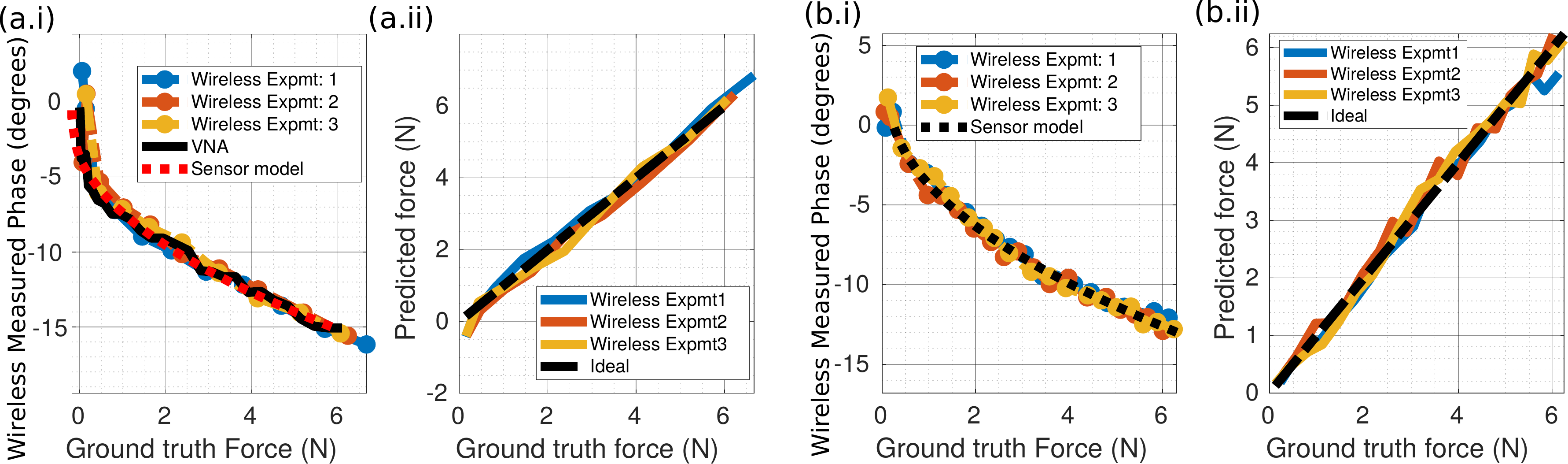}
    \caption{(a.i) Shows the wireless measured phase changes from the rigid PCB integration and (b.i) shows the same for standalone PCB free integration with commercial RFID stickers. (a.ii) and (b.ii) plots the predicted forces from the sensor model for both these sensors and shows that the predicted force is close to the estimated force, with ideal 0 error curve being when predicted and estimated forces become equal to each other (black dash line)}
    \label{fig:rfic_pcb}
\end{figure}

To evaluate our sensor prototypes, we repeat the same experiment of applying increasing levels of forces and measuring phases as discussed in Section~\ref{sec:implementation} for our wired experiments. However, this time, we use the rigid PCB with matching network and RFID IC connected to a dipole antenna (Fig.~\ref{fig:exp_setup} b.i) which allows to obtain wireless measurements from the RFID reader.
The RFID reader collects phase information as the indenter applies forces on the sensor, and implements the phase diff algorithm as in Section~\ref{sec:design} to obtain phase differences which can be mapped to applied forces.
The designed rigid PCB also allows to depopulate the matching network to disconnect the sensor from the RFID IC, and then by connecting it via the UFL cable to VNA, we can obtain the ground truth phase readings from the VNA.
Hence, in addition to wireless phase measurements, we can also collect wired phases from the VNA, and as seen in Fig.~\ref{fig:rfic_pcb}a.i. Both the ground truth wired phase measurements, as well as wireless measurements across multiple experiments (three experiments shown for brevity) are consistent, which shows that we can indeed sense phases wirelessly with a good accuracy.
We also model the sensor using a 2nd order polynomial fit based on the collected VNA data and use it to obtain force from wireless phase measurements. We illustrate the closeness of the predicted force from wireless measurements and ground truth readings from the load cell in Fig.~\ref{fig:rfic_pcb}a.ii.

Moving ahead, we repeat the same experiment but this time with standalone integration of the sensor with RFID sticker (Fig.~\ref{fig:exp_setup}b.ii).
A major difference here is that the sensor has two tungsten filament instead of a single one for the PCB-based integration.
We observe similar results to the previous experiments (Fig. \ref{fig:rfic_pcb} b.i, b.ii), which shows the flexibility in which our sensor can be integrated to existing RFIDs, using a single tungsten wire filament on a PCB, or using two tungsten wire filaments and sticking it to the RFID antenna pads.

\begin{figure}[t]
    \centering
    \includegraphics[width=\textwidth]{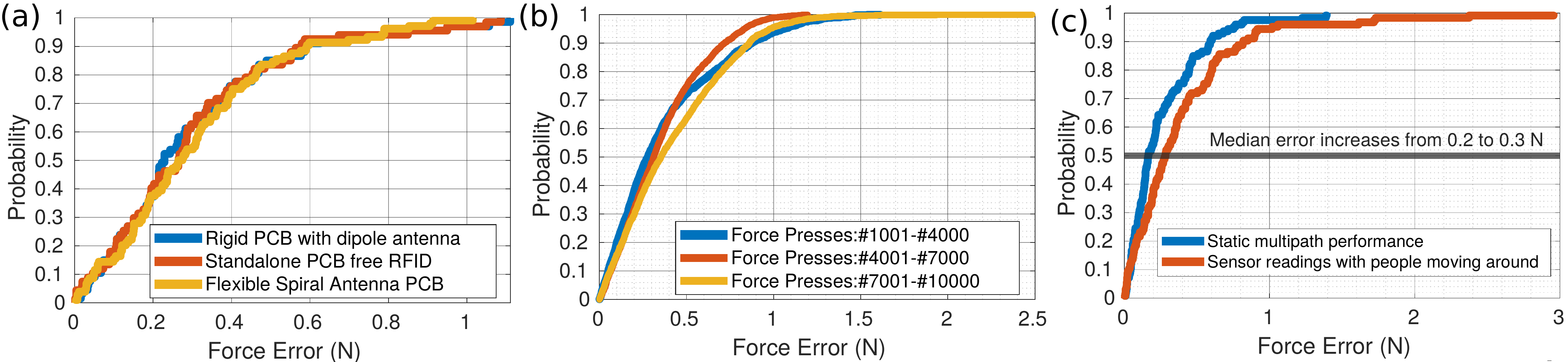}
    \caption{Benchmark studies: (a) Setup image with spiral antenna (a) Both PCB and standalone and small flex PCB integration give similar performance (c) Sensor durability}
    \label{fig:benchmarks}
\end{figure}

We also repeat the same experiment with the spiral antenna flexible PCB to obtain similar performance, as visible in Fig.~\ref{fig:exp_setup}b.iii (the phase to force curves are not shown for brevity sake). We compare the results of these three experiments by calculating the force error with ground truth and plotting a CDF (Fig.~\ref{fig:benchmarks}a), which clearly shows that the three sensor prototypes show similar median errors of about $0.2$-$0.3$~N, which is by itself about 2 times better than previously shown results for backscatter based force sensor at $900$~MHz (\cite{wiforce} which had $0.6$~N error at $900$~MHz).

We also stress test the spiral antenna PCB and apply more than $10,000$ force presses on the sensor to see the durability of the sensor (Fig.~\ref{fig:benchmarks}b). We observe that the median error increases only slightly during the course of multiple trials on the sensor.
One main reason why \name's capacitive sensor is durable is that it doesn't have wired interconnects with capacitor which have known to hamper the durability of such sensors~\cite{reskin}.
A wireless capacitive sensor doesn't suffer from these drawbacks, and thus as compared to previous capacitive-based sensors, \name is much more durable as a consequence of the wireless operation.

\begin{figure}[t]
    \centering
    \includegraphics[width=\textwidth]{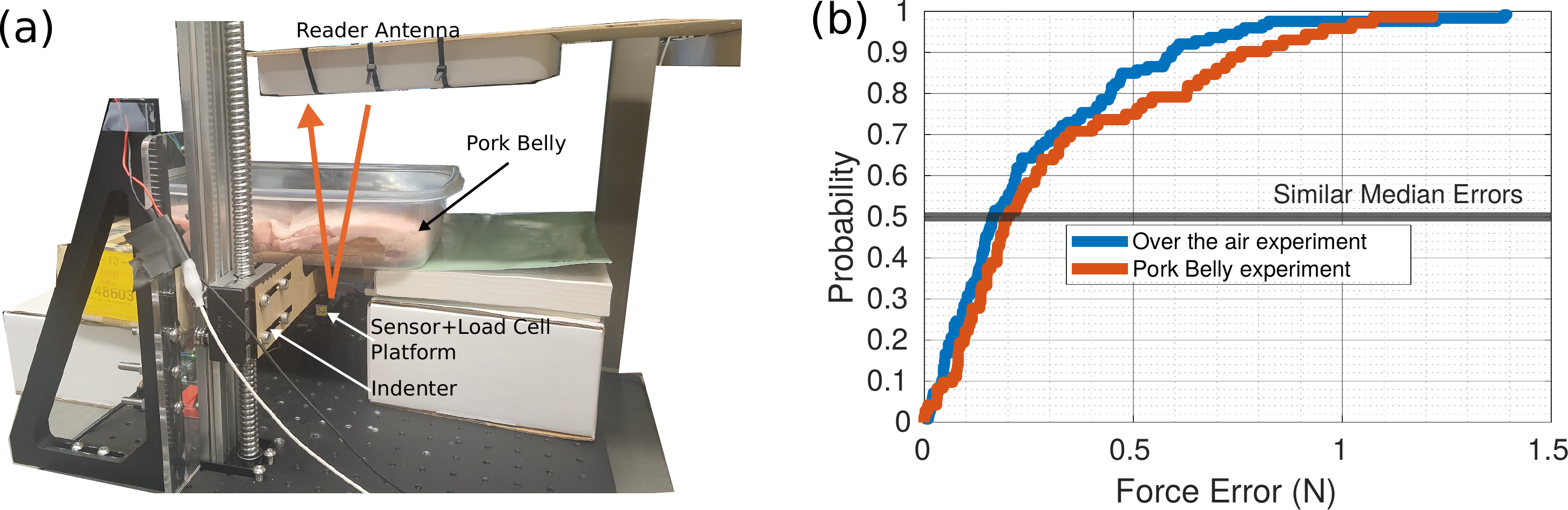}
    \caption{(a) Shows the test setup with sensor load cell platform occluded via the pork belly, and the RFID reader antenna is placed directly over the pork belly to ensure propagation occurs via the tissues and not over the air (orange arrow) (b) Shows that having propagation over the pork belly setup doesn't adversely affect \name's sensing results}
    \label{fig:multipath_pork}
\end{figure}

One problem with RFID based sensing systems relying on phase transduction is that these systems are not robust towards dynamic multipath. The reason is that the dynamic multipath also produces phase shifts which can be confused for force measurements.
However, these movements usually cause erratic phase shifts which can be handled by averaging across the multiple RFID channels which would see a different phase shift due to multipath but a consistent phase shift due to applied forces.
Since \name's phase processing algorithm handles consistent averaging across multiple RFID channels, our sensing strategy is robust to dynamic multipath since the median error increases only slightly (from $0.2$ to $0.3$~N) when we have people moving around near the experimental setup (Fig.~\ref{fig:benchmarks}c).

We also tweak our experimental setting by occluding the experimental setup with a $10$~cm thick pork belly layer (Fig.~\ref{fig:multipath_pork}a), in order to verify if the sensor can work for in-vivo applications. Since UHF RFIDs use $900$~MHz frequency which is $<1$~GHz, it is a good frequency for tissue propagation~\cite{wiforce,vasisht2018body,gupta2003towards,dove2014analysis} and hence there are very minimal changes in sensor performance (Fig.~\ref{fig:multipath_pork}b) when we introduce the pork belly to the experiment setup. 
We keep the RFID reader antenna closer to the pork belly setup to ensure there is no signal path over the air between the sensor and RFID reader antenna and the propagation happens solely via the pork belly.
These results are consistent with past work~\cite{wiforce} also showing that using $900$~MHz frequencies allow for similar performances over the air and beneath tissues.

\begin{figure}[h]
    \centering
    \includegraphics[width=\textwidth]{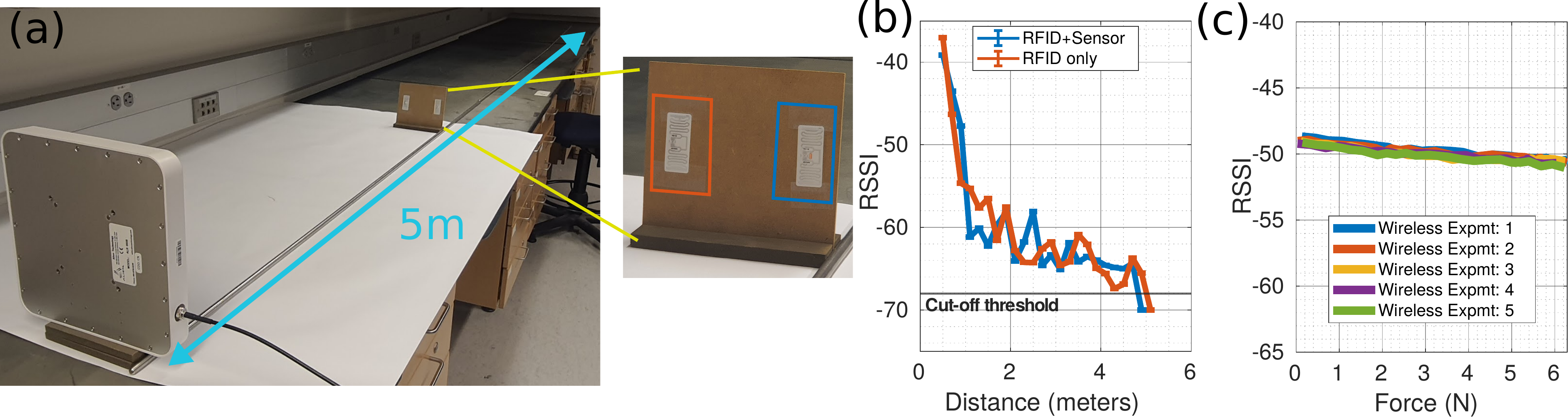}
    \caption{(a) We move the cardboard plate away in steps of 20cm from 0.5m to 5m distance from the RFID reader antenna. The cardboard plate has two RFIDs, one vanilla RFID (orange box) and one RFID with the force sensor interfaced (blue box) (b) Both the blue and orange RFIDs show similar range results (c) Sensor RSSI doesn't degrade by much because of Force capacitance change, unlike phases. Hence (b) which has sensor at zero force and (c) showing no dependence of sensor RSSI on force together show that \name's sensor interfacing doesn't adversely affect the RFID range}
    \label{fig:rfid_rssi}
\end{figure}

To conclude the sensor benchmarks, we evaluate the range of \name sensor, and show that integration of \name into existing RFIDs does not drastically affect the reading ranges. For this experiment, we have a normal RFID without the sensor and a RFID created with the sensor side by side and we take the signal strength readings from $0-5$~m from the RFID reader antenna (Fig.~\ref{fig:rfid_rssi}a) and plot the readings side by side (Fig.~\ref{fig:rfid_rssi}b). 
We observe that the RFID with sensor fails to be read beyond $4.6$~m, whereas the standard RFID goes till $4.8$~m, which is only minimally more than sensor and RFID.
The reason for this is the fact that the sensor is purely capacitive and mostly produces change in signal phase when force is applied, and does not change the signal amplitude by much. 
This is also shown when we plot the RSSI vs force for the sensor, placed at about $1$~m away, and see that unlike phase readings which show changes as force is applied, the RSSI readings stay more or less constant (Fig.~\ref{fig:rfid_rssi}c), with only a dip of about $-1$~dB (and hence the minimal range reduction of $0.2$~m).

\subsection{Case Study 1: Knee Force Measurement}

\begin{figure}[t]
    \centering
    \includegraphics[width=\textwidth]{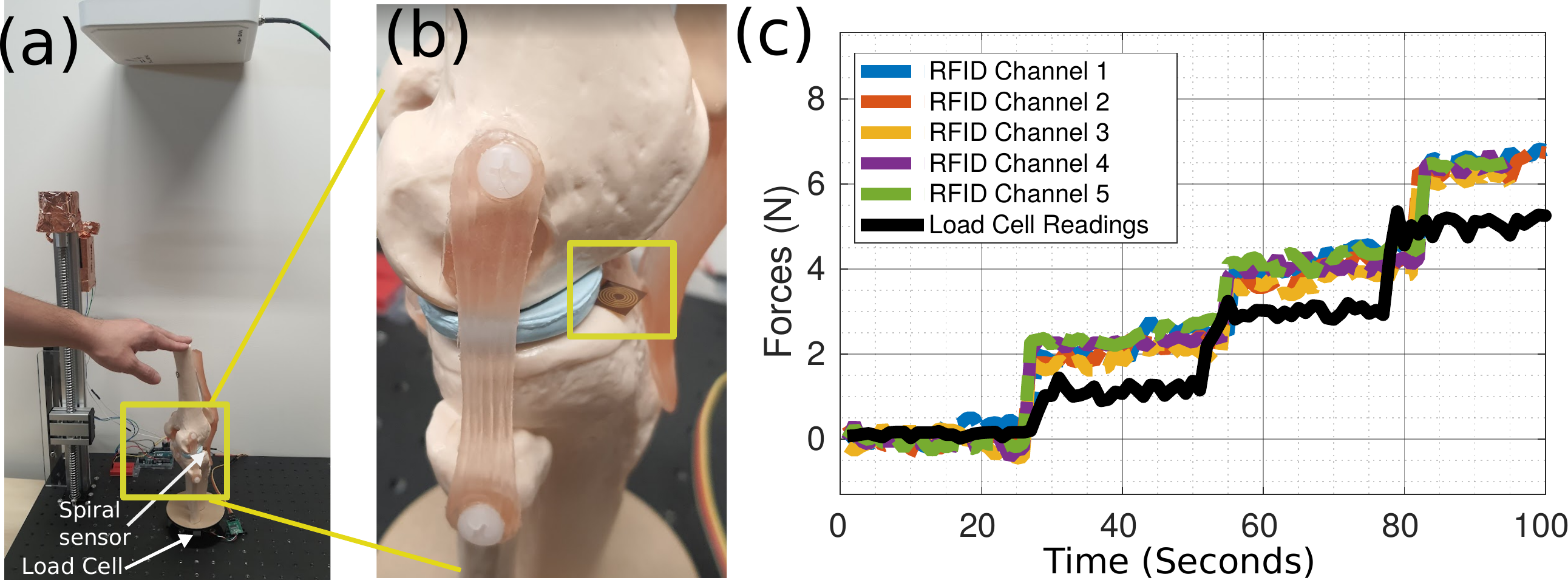}
    \caption{(a) Test Setup (b) Zoomed in image to show the sensor and (c) The sensor is able to read the forces in a wireless batteryless manner, which correlates well with the load cell readings}
    \label{fig:knee_res}
\end{figure}

For the first case study, we place the spiral antenna flexible PCB beneath the knee joint (Fig.~\ref{fig:knee_res}a,b) in a toy-knee model and apply forces onto the sensor by pressing on the top of the knee bone manually. Our toy-knee represents a prosthesis that could wear other time, and for which measuring the level of force transmitted gives indication on the wear level.
As seen clearly the small size form factor of \name sensor fits inside the knee joint without disturbing the joint at all and is almost un-noticeable if viewed from a distance.
To see if the sensor is able to detect the applied forces, we apply roughly $1$~N, $3$~N, $5$~N force in steps delayed by $20$~sec.
This is made possible by showing the applied forces on a GUI via the load cell so that we can manually adapt the the applied load such that approximately this much force is applied on the sensor (see Fig.~\ref{fig:knee_res}).
The load cell is compensated to zero at the knee-model's weight to compensate for the static gravitational force.
Because of this compensation, the actual applied forces on the sensor is the applied force by hand + static weight of the top part of the knee model relatively to the joint (approximately $1$~N) and hence the readings reported by the sensor are $1$~N offsetted when the sensor is manually pressed.
The phases collected (and hence the estimated forces) show repeatable behaviour across the various channels and show three clean jumps corresponding to when the sensor is pressed with higher forces (Fig. \ref{fig:knee_res}c), which shows that the sensor was indeed able to capture the applied forces via the knee joints.
This motivates the use of \name sensors to create smart implants, which can sense the applied forces and can be then used to monitor the implant health.

One point to note here is that via the toy-knee model we can get forces in range of $0$-$6$~N, however in an actual knee the forces would be much higher (say for a person weighing $60$~Kg, the force on the joint could be as high as $300$~N). So, in actual usecase the sensor has to be re-designed for higher forces and we present simulation results to show how this is possible later in this section. However, this case study still showcases two key results, one, the sensor can fit into these tight constraint spaces and two, the sensor can read forces applied unevenly by the joints in addition to the previous results shown when sensor was pressed by an actuated indenter.

\subsection{Case Study 2: Box Item Weighing}

\begin{figure}[t]
    \centering
    \includegraphics[width=\textwidth]{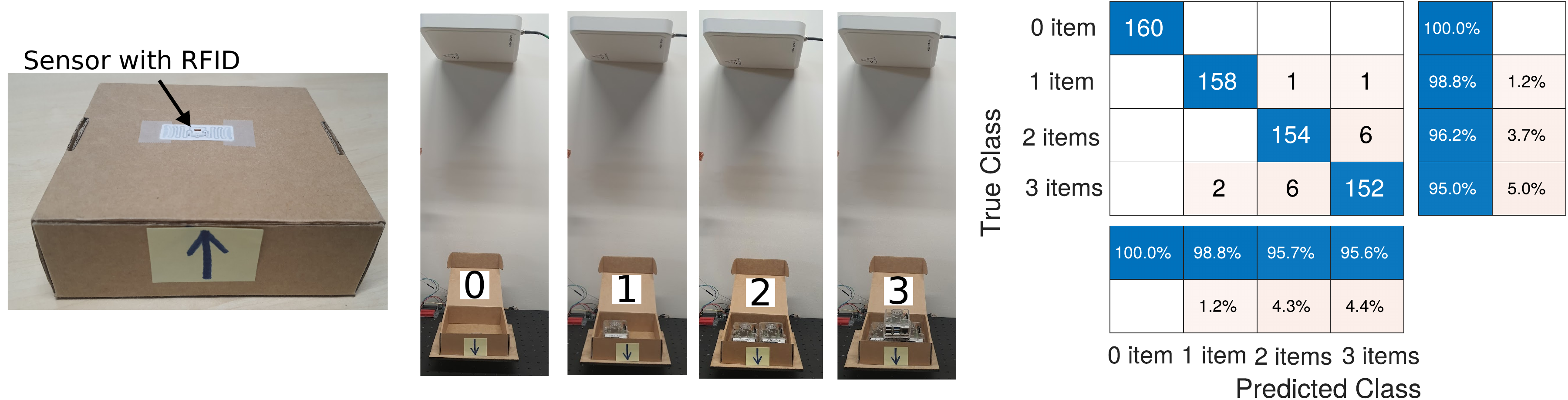}
    \caption{Weighing objects placed in a box with \name RFID sticker stuck to the bottom of the package: By measuring the weight of the contents, \name is able to classify the number of contents in the box}
    \label{fig:box_res}
\end{figure}

Now, for the second case study, we stick a RFID sticker interfaced with a sensor to the bottom of a cardboard package.
We place 3 items in the box one at a time, and record the phase shifts from the sensor (Fig.~\ref{fig:box_res}).
For the experiment, we choose the item as an Raspberry Pi 4 with a plastic case which weighs about $200$~g and hence applies $2$~N force on the sensor.
Hence, when 1 item is placed the sensor would be under $2$~N force, which then increases to $4$~N and finally $6$~N when three items are placed.
The reader is able to identify the number of items placed in the box since when more items are kept in the box, more gravitational force is applied to the sensor and hence there is a different associated phase shift when the items are placed.
We take $160$ measurements of placing items in the box and classify the number of items based on the measured phase difference (and hence the force), to plot the confusion matrix as shown in Fig.~\ref{fig:box_res}. 
The sensor is able to detect the classes with $> 95$\% accuracy, with the higher forces being slightly confused with lower ones, because the force to phase profile is not exactly linear and the phase difference between $4$~N and $6$~N is lesser than the difference between $0$~N, $2$~N and $2$~N, $4$~N.
However, since even the 95\% errors of \name as seen from CDFs before are about $1$~N, these results are consistent with the CDF profiling of force readings.

\subsection{Generalization to other force ranges}
\begin{figure}[t]
    \centering
    \includegraphics[width=\textwidth]{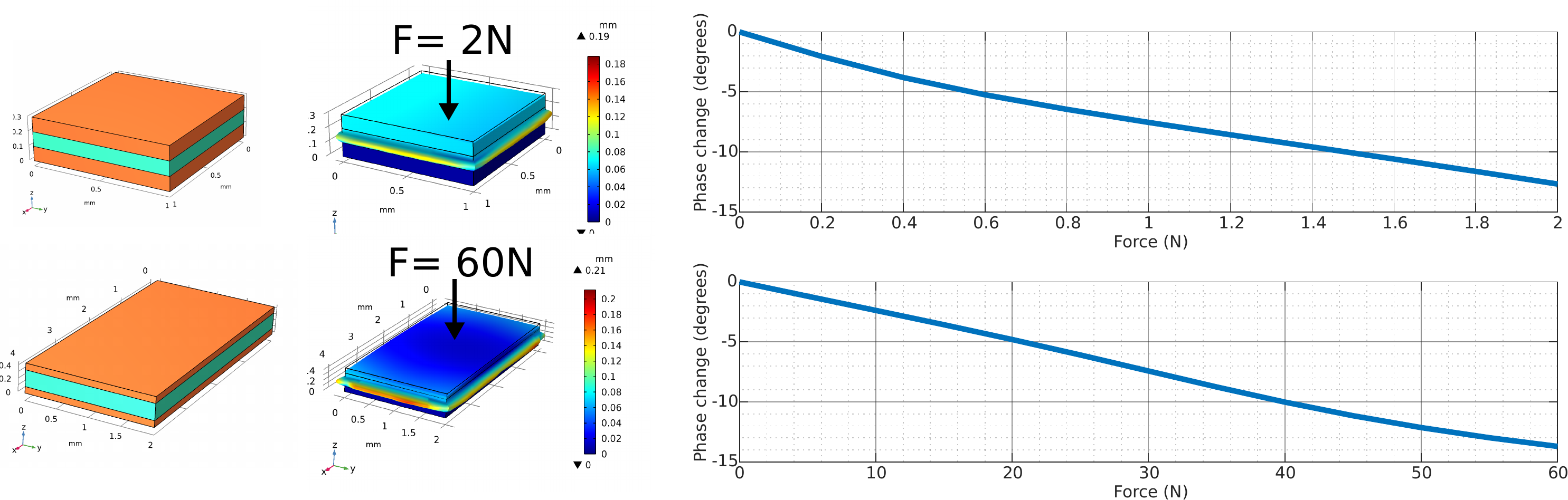}
    \caption{Simulations to show different sensor designs for lower and higher force magnitudes}
    \label{fig:gen_sim}
\end{figure}

\name designs a force sensor which works for ($0-6$)~N range. However the design techniques discussed are fairly general and can be used to design a custom force sensor to fit the diverse applications as required.
Further, our multiphysics simulation framework allows for tweaking the sensor dimensions appropriately, and applying different force magnitudes to characterize the working of sensors in different form factors.
To show this, we present simulation results with other sensor geometries ($1$~mm~$\times$~$1$~mm~$\times$~$0.3$~mm sensor with Ecoflex 00-30 customized to sense 0-2N) and soft materials as dielectrics (same dimensions $4$~mm~$\times$~$2$~mm~$\times$~$0.4$~mm sensor but with stiffer neoprene rubber polymer than Ecoflex 00-30).
These sensor designs give similar phase change results (about $15^{o}$) for both lower ($0-2$)~N as well as higher ($0-60$)~N force ranges (Fig.~\ref{fig:gen_sim}). Hence, one can use the COMSOL Multiphysics simulations to first come up with the correct sensor geometry and material choice to match the given force range, fabricate the sensor via the given steps and then integrate it with RFIDs as appropriate, either with a flexible PCB or existing RFID stickers.

\section{Discussion and Future Work}\label{sec:conlusion}

\subsection{Reading forces from multiple sensors concurrently}
In \name evaluations, we have read from one RFID sensor tag at a time. However, RFID MAC has advanced significantly over the years, and easily generalises to multiple tag readings via the same RFID reader, and then by observing the channel estimates we can estimate the applied forces on each of the tag.
Thus, \name sensing principle extends very naturally to multiple such force sensors.
Hence, \name can easily demonstrate this capability to read from multiple tags concurrently and motivate new applications where reading from large number of tags simultaneously would be beneficial.

\subsection{Integration with other backscatter technologies}
\name also shows how we can design such analog backscatter sensors and integrate them to existing RFID ICs by having the sensor as a PCB pad component.
This feature of \name's capacitive sensor allows possible integration with upcoming backscatter technologies as well, like Wi-Fi~\cite{hitchhike,freerider}, LoRA\cite{guo2022saiyan,talla2017lora} and UWB backscatter\cite{pannuto2018slocalization}.
Unlike UHF RFID, these backscatter technologies are still under research with no commercially available IC to interface the sensor in the present day.
But, in the future, such analog capacitive force sensors can be integrated with these new technologies using similar methods like PCB interfacing of the sensor and standalone method. 

\subsection{Biocompatible hermetic packaging of the sensor}
Although we fabricate \name sensor on flexible sticker-like RFIDs and PCBs, and also show them working under pork belly and beneath toy-knee models, another important step is to cover the whole sensor in  a biocompatible packing which is also known as hermetic packing, which refers to airtight moisturetight packing.
There have been previous examples of hermetically packed batteryless sensors used in-vivo, for example wireless monitoring of blood pressure near the arteries~\cite{fonseca2005implantable,issa2019cardiomems} and non-invasive EMG recording~\cite{seo2016wireless} which use materials like nitinol and medical grade epoxy to encapsulate the sensor and pack it hermetically.
In the future, \name can also adopt similar methods and encapsulate the sensor to make it bio-compatible.

\subsection{Force sensing with sensor under movement}
In \name evaluations the sensor is kept fixed and not moved when forces are applied onto it.
However, for applications like safe control of robot via force feedback, the robot would move around and hence the sensor attached to the robot would not be stationary.
Although we have shown that \name's sensing strategy is robust to movement of environmental entities after averaging across the RFID channels, when the sensor itself starts to move simple averaging won't help to compensate for movement induced phase changes.
These mechanical movements would show up as periodic, or having a certain spectral signature when taking a frequency transform of the phase readings as shown by past RFID sensing and localization works~\cite{li2016paperid} which also characterize phase changes as RFIDs move around.
Hence, in the future, we can also write a frequency transform to remove the movement artefacts and make the \name sensors robust to sensor movements.


\section{Related Works}
\name presents the first batteryless sticker-like form factor force sensors which enable numerous applications like ubiquitous weight sensing and in-vivo applications like knee-implant force sensing. In this section we will compare the past work on force sensors and RFID based sensing and put them in context with \name's contributions.

\subsection{Discrete MEMS Force Sensors}

Force sensors has been an active area of research over the past years in the flexible electronics and MEMS sensor communities~\cite{chi2018recent,dahiya2013directions,soni2020soft,rosenberg2009unmousepad,zou2017novel,parzer2016flextiles,parzer2018resi,pointner2020knitted,lee2017durable,piezo2}.
Today's force sensors can be roughly divided into three groups, impedance based sensors, piezo sensors and magnetic sensors~\cite{piezo1,piezocatheter,piezo3,resistive1,resistive2,capacitive1,capacitive2,capacitivesurvey,reskin,magnetic2}.
The impedance based sensors usually transduce force onto changes in sensor resistance~\cite{resistive1,resistive2,rosenberg2009unmousepad} or capacitance~\cite{capacitivesurvey,capacitivetina,capacitive3}.
The piezo sensors generate small voltages as a function of contact forces applied onto the sensor~\cite{piezo1,piezocatheter,piezo3,ramesh2021sensnake}.
The magnetic sensors create distortions in magnetic field due to the applied force~\cite{reskin,magnetic2,magnetic3,magnetic4}.
A unifying theme of these three sensor types is the fact that they all need to be digitized before communicating these readings to remotely located wireless reader.
This is because the effect generated by force can not be read directly at a considerable distance, as ultimately these sensors map to either voltage or magnetic field fluctuations which die out within few centimeters.
Typically, past work addresses this by using amplifiers and digitization blocks in Data acquisition units (DAQs) for the sensors. 
Usually, a DAQ would consist of OPAMP amplifier circuits, ADCs for resistive and piezo sensors~\cite{ibrahim2018experimental,dahiya2011towards}, CDCs for capacitive sensors~\cite{capacitivetina,kim2015force}, and magnetometers for magnetic sensors~\cite{reskin,magnetic3}.
After digitization, the values are modulated onto a wireless signal to communicate the readings to a distance.
However, this process of digitization, and further modulation usually requires dedicated electronics typically in form of low-power embedded microcontrollers which do not allow these MEMS sensors to attain the batteryless and sticker like form factor for wireless force feedback.

\subsection{Joint communication and sensing based sensors}
\name sensor shows how force can be transduced to wireless signal phases directly, without requiring these extra digitization+modulation steps. 
Thus, the sensors which enable a similar wireless transduction form the closest related works to \name, however none of them can be read as robustly, and have the sticker-like form factor like \name.
A common set of works utilize inductor coils with capacitive sensors to form LC circuits~\cite{li2015review, huang2016lc,boutry2019biodegradable,nie2018droplet}.
LC circuits have the property of absorbing a certain resonant frequency depending on the value of capacitance, and thus as the force applied changes capacitance, the force information gets transduced onto the resonant frequency absorbed.
However, these sensors have shown limited range (few cm), and unreliable reading in cluttered environments, and less dynamic range as the resonant frequency shifts only by a few kHz/MHz~\cite{bau2018contactless,huang2016lc}.
Further, these sensors do not generalize easily to multiple sensors since there is no way to determine the identity of certain sensor from the resonant frequency alone.

Similar to LC sensors, we have strain sensors which can be SAW based~\cite{li2018miniature,yi2011passive,thai2011design,humphries2012passive}, or even some RFID based strain sensors~\cite{teng2019soft}.
These strain sensors don't really sense the contact force but are capable of sensing the shear forces which create elongations.
These elongations change the physical dimensions of the RFID antenna, or the SAW channel filter, which again leads to changes in the antenna's resonant frequency and thus can be sensed similar to LC sensors, and have similar drawbacks on not generalizing to a variety of environments since other objects in environment also absorb certain frequencies~\cite{yi2011passive}.
Another prior work~\cite{wiforce} also attempts to map force to phase, however as discussed earlier does not achieve the sticker form factor and the designed prototype is not batteryless since it is not evaluated with an energy harvester.


\subsection{RFID based sensing of temperature, light, touch gestures and other quantities}

In addition to design of a mm-scale force sensor, we have also shown how to integrate the sensor with RFID systems.
There has been a vast body of works on how to use RFIDs to sense effects like environmental temperature~\cite{pradhan2020rtsense,chen2021thermotag}, moisture~\cite{hasan2015towards,wang2020soil}, photointensity~\cite{omidRFID,amin2014chipless}, binary touch/no-touch~\cite{hsieh2019rftouchpads,li2016paperid,pradhan2017rio} etc.
One particular past work showcases a generalized framework for interfacing various sensors with RFIDs~\cite{omidRFID}, however this method ends up compromising the RFID read range since the main method of sensing proposed is differential wake up thresholds.
This ends up hampering the sensor resolution as well since the amplitude of reflected signals decreases.
Also~\cite{omidRFID} just motivates the possibility of force (pressure) sensor integration using the proposed system, however does not explicitly evaluate it like how temperature and photointensity were evaluated. 
In \name we show how to integrate force sensors without affecting the range of the RFID, by doing a purely phase based transduction, and even integrating with common commerically available RFID tags.

\noindent Unlike the other sensed quantities, force sensing has not been extensively studied for RFID sensing systems. There has been just a single work~\cite{fernandez2015passive} which used a particular SL900A RFID tag~\cite{AMSSL900A} that exposes an ADC interface to allow connecting various sensors, and the authors of~\cite{fernandez2015passive} connect up a force sensitive resistor to the ADC pin.
However, this interfacing of the resistor to the ADC pin requires extra accompanying electronics to enable sensor data buffering and data register writing which put a massive burden on energy harvester since RFIDs operate on a very tight energy budget.
In comparison, \name allows for force sensors to be integrated in analog domain with RFIDs without putting extra burden on the digital communication, and allowing for integration with any commercially available tag today, not just a specialized tag with ADC pins.


Finger touch sensing is another popular application of RFID sensing systems~\cite{marquardt2010rethinking,sample2009capacitive,simon2014adding,schmidt2000enabling,gao2018livetag,pradhan2017rio,li2016paperid,li2015idsense}, which is related to force sensing however usually these systems sense gestures and binary contact/no-contact. 
IDSense~\cite{li2015idsense}, PaperId~\cite{li2016paperid} and RIO~\cite{pradhan2017rio} are some notable works which demonstrate relationship between phases and finger touches, simple manufacturing of such touch sensitive tags as well as multi-RFID touch generalization. However, these past work are oblivious to different force levels applied during the touch process, and these works just sense the places where the tag is interrogated via the fingers to sense simple gestures/sliding movements etc.


\bibliographystyle{unsrt}
\bibliography{paper}

\end{document}